\documentclass[10pt,twocolumn,letterpaper]{article}

\usepackage[pagenumbers]{cvpr} %

\usepackage{graphicx}
\usepackage{amsmath}
\usepackage{amssymb}
\usepackage{booktabs}
\usepackage{xcolor}
\usepackage{colortbl}
\usepackage{enumitem,eucal}
\usepackage{stfloats}  %
\usepackage[ruled, lined, linesnumbered, commentsnumbered, longend]{algorithm2e}

\graphicspath{{../../../}{../../}{../}{./latex/figures/}{latex/figures/}}

\usepackage[pagebackref,breaklinks,colorlinks]{hyperref}

\usepackage{caption}
\usepackage[capitalize]{cleveref}
\crefname{section}{Sec.}{Secs.}
\Crefname{section}{Section}{Sections}
\Crefname{table}{Table}{Tables}
\crefname{table}{Tab.}{Tabs.}

\definecolor{GRAY}{gray}{.9}
\definecolor{LGRAY}{gray}{.94}

\def\confName{CVPR}
\def\confYear{2022}

\def\ourmethod{{RMat}\xspace}  %

\begin{document}

\title{Boosting Robustness of Image Matting with Context Assembling and \\
Strong Data Augmentation
}

\author{
Yutong Dai$^1$, ~~~~~~~
Brian Price$^2$, ~~~~~~~
He Zhang$^2$, ~~~~~~~ 
Chunhua Shen$^3$\\[0.2cm]
$^1$The University of Adelaide
~~~~~
$^2$Adobe Inc. 
~~~~~
$^3$Zhejiang University 
}

\twocolumn[{%
\renewcommand\twocolumn[1][]{#1}%
\maketitle
\begin{center}
    \centering
    \captionsetup{type=figure}
    \includegraphics[width=\textwidth]{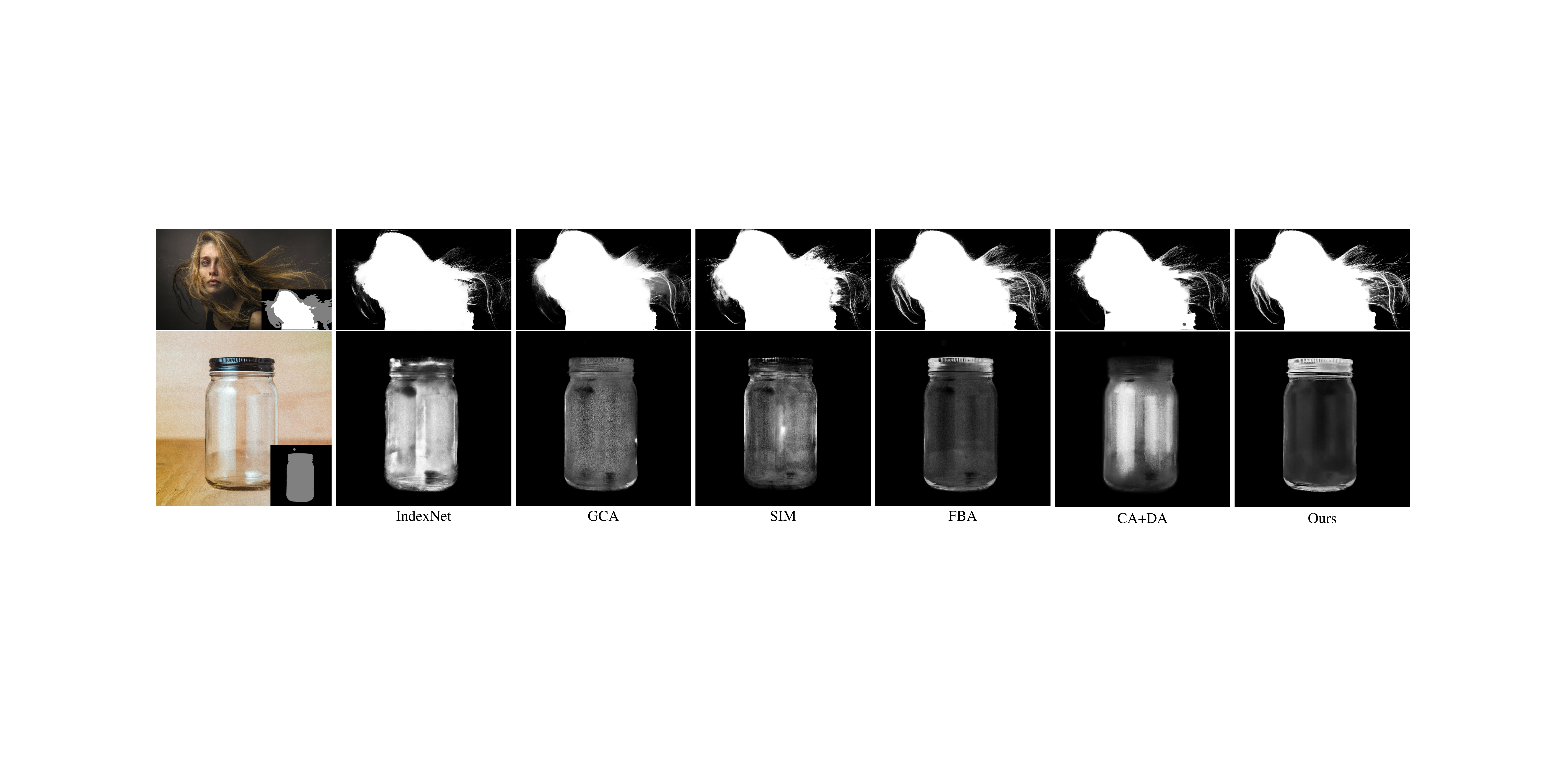}
    \vspace{-10pt}
    \captionof{figure}{Matting results on real-world images. From the second column to right are results of IndexNet~\cite{lu2019indices}, GCA~\cite{li2020natural}, SIM~\cite{sun2021semantic}, FBA~\cite{forte2020f}, CA+data augmentation ($\mathcal{DA}$)~\cite{hou2019context} and our method, respectively. Note that, all the methods are trained with the DIM~\cite{xu2017deep} dataset (except SIM is trained with the SIMD~\cite{sun2021semantic} dataset). They are comparable on benchmark images, while presents varying results on real-world images. Our method shows better generalization ability. 
    }
    \label{fig:example}
\end{center}%
}]
\vspace{5pt}

\begin{abstract}
   Deep image matting methods have achieved increasingly better results on benchmarks (e.g., Co\-m\-po\-si\-ti\-on-1k$/\tt al\-ph\-a\-ma\-tting.com$). However, the robustness, including robustness to trimaps and generalization to images from different domains, is still under-explored. 
   Although 
   some works propose to either refine the trimaps or adapt the algorithms to real-world images via extra data augmentation, none of them 
   has 
   taken both into consideration, %
   not to mention the 
   significant performance deterioration on benchmarks while using those data augmentation. To fill this gap, we propose an image matting method which achieves higher robustness (\ourmethod) via multilevel context assembling and strong data augmentation targeting matting. Specifically, we first build a strong matting framework by modeling ample global information with transformer blocks in the encoder, and focusing on details in combination with convolution layers as well as a low-level feature assembling attention block in the decoder. Then, based on this strong baseline, we analyze current data augmentation and explore simple but effective strong data augmentation to boost the baseline model and contribute a more generalizable matting method. Compared with previous methods, the proposed method not only achieves state-of-the-art results on the Composition-1k benchmark ($11\%$ improvement on SAD and $27\%$ improvement on Grad) with smaller model size, but also shows more robust generalization results on other benchmarks, on real-world images, and also on varying coarse-to-fine trimaps 
   with our extensive experiments.%
\footnote{This work was in part done when YD was an intern at Adobe and CS was with The University of Adelaide. CS is the corresponding author.}
\end{abstract}

\section{Introduction}
\label{sec:intro}
Image matting, as a fundamental computer vision task, aims to obtain the high-quality alpha matte of the foreground object given an input image. Mathematically,  image matting is  formulated as:
\begin{equation}\label{matting_eq}
    I_i=\alpha_{i}F_i+(1-\alpha_{i})B_i \,,
\end{equation}
where $I_i$, $\alpha_i$, $F_i$ and $B_i$ are observed colour value, alpha value, foreground value and background value of pixel $i$, respectively. It is an ill-posed problem because there are $7$ unknowns given $3$ equations. With recent success of deep learning, deep matting methods~\cite{xu2017deep,lu2019indices,hou2019context,li2020natural,dai2021learning,sun2021semantic,forte2020f} achieve promising results on benchmarks such as Composition-1k~\cite{xu2017deep} and $\tt alphamatting.com$~\cite{rhemann2009perceptually}. While increasingly higher accuracy have been promised on benchmarks, due to the limited training/test data, robustness of these methods
is still under explored.

\begin{figure*}
    \centering
    \includegraphics[width=0.95\linewidth]{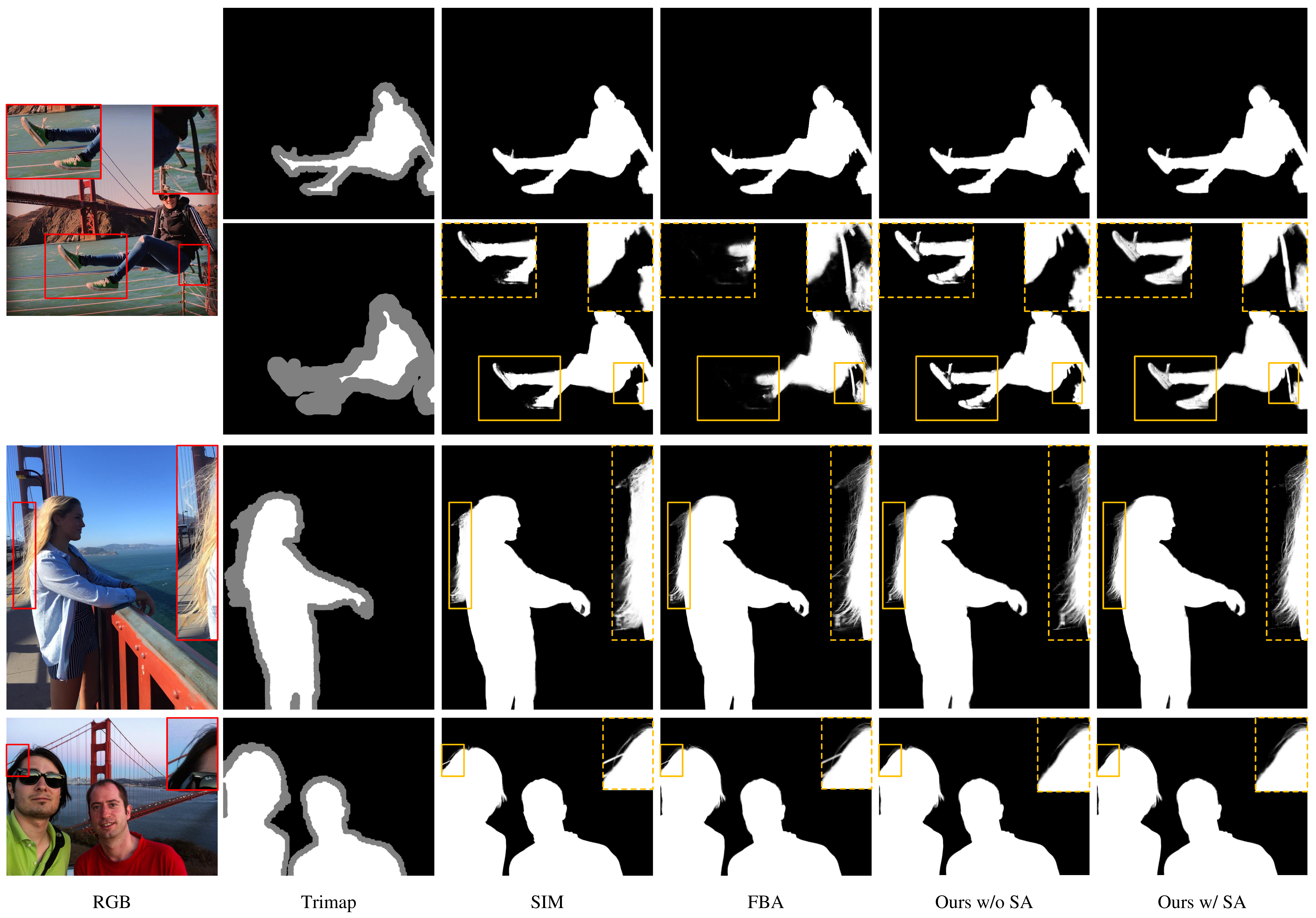}
    \caption{Visual results on real-world images showing robustness of methods. The methods in comparison are SIM~\cite{sun2021semantic}, FBA~\cite{forte2020f}, and our method w/o and w/ Strong Augmentation ($\mathcal{SA}$), respectively. The first two rows are results of the same RGB image with different trimaps. Our methods are more robust to various trimap precision. In the third row, our model using $\mathcal{SA}$ captures better details. The last row presents the benefit of modeling global context: the bridge component on the left top side is apart from the main body of the bridge, which is recognized as foreground in SIM and FBA. Our method, however, distinguishes it from the foreground human clearly thanks to the context assembling. Best viewed by zooming in.}
    \vspace{-10pt}
    \label{fig:real_images}
\end{figure*}

First, robustness to the trimap is important for a matting algorithm. In real applications, trimaps are labeled by users, with unpredictable precision of unknown regions.  However, as shown in  Fig.~\ref{fig:real_images},  existing matting methods \cite{sun2021semantic,forte2020f} are sensitive to the shape/size of the given trimap so that it requires users' more time to accurately brush the trimap. A main reason why existing methods are sensitive to the precision of trimap is they focus more on detailed cues, where robustness to trimap with varing precision,  which relies more on context information, is less cared about. One possible solution is to optimize the trimap to be a more detailed one.
This was proposed in~\cite{cai2019disentangled}, where an extra branch was used to generate a more precise trimap. Though multi-task learning is leveraged in this method to adapt the trimap, its context modeling is still limited, which restricts its robustness in applications.
Therefore, we wonder \textit{ whether  it is possible to enhance the context modeling ability (robustness) of a  matting algorithm with a simpler and more effective approach}.

Meanwhile, it has been known that deep matting models trained on synthetic data undertake the risk of poor generation to real-world  domains~\cite{hou2019context,sun2021semantic,yu2021mask} (Fig.~\ref{fig:example}). However, due to the difficulty of obtaining ground-truth alpha matte annotations for real-world images, only synthetic datasets are available to train the matting algorithms, 
so some works 
attempted 
to narrow the domain gap.
For example, 
\cite{hou2019context,yu2021mask} leverage extra data augmentation to adapt the models to real-world images, while significant performance degradation on the synthetic benchmark happens at the same time. 
Although better prediction on real-world images is appreciated, it is desirable that 
the model %
can be 
generalized to broader scenes without sacrificing too much performance on 
images from one domain such as the benchmark data, 
because it is hard to confirm which domain a test image comes from, 
and not to mention that the real-world test images in~\cite{hou2019context,yu2021mask} can only cover a tiny part of real scenes. Therefore, \textit{a model showing better domain generalization ability is in demand}.

Motivated by these demands, we present a more robust matting method (\ourmethod), which achieves higher robustness to diverse trimap precision and better generalization to various domains. In detail, two steps are %
designed. 
The first step is to %
build
a strong baseline model with multilevel context assembling. It is implemented by combining transformer blocks with convolution layers, where global context is learned via self-attention modules and local context is emphasized by convolution layers. 
Considering the uniqueness of matting that needs local context information 
and original test resolutions
to capture details, we explore designs and implementations aiming at this task to build an efficient model. Further, founded on this strong baseline model, we investigate strong data augmentation for matting. 
We analyze the problems behind current augmentation and propose strong augmentation strategies specifically for matting.
Finally, to verify robustness of the model, a series of experiments and visualizations are carried out in comparison with state-of-the-art methods.

In summary, our main contributions are:
\textbf{1)} A strong matting framework with multilevel context assembling;
\textbf{2)} Strong augmentation strategies targeting matting;
\textbf{3)} Designs of experiments and visualizations to verify generalization capability of matting models; 
\textbf{4) }State-of-the-art results on benchmarks (w/ and w/o fitting the training sets), higher robustness to varying trimap precision, and better generalization to real-world images.

\section{Related Work}
\label{sec: related}
\textbf{Deep Image Matting.}
Before the success of deep learning, conventional matting methods~\cite{levin2007closed,chen2013knn,wang2007optimized,he2011global,shahrian2013improving,chuang2001bayesian,sun2004poisson} dominated this field by solving Equation~\eqref{matting_eq} using different assumptions such as the local smoothness assumption in~\cite{levin2007closed}. Due to the nature of relying on low-level color cues, their assumptions are easily violated in complex images. To overcome this dilemma, deep matting methods~\cite{xu2017deep,tang2019learning,lu2019indices,hou2019context,cai2019disentangled,li2020natural,dai2021learning,sun2021semantic,forte2020f,lin2021real,yu2020high} appeared with the development of deep learning. 

Among recent state-of-the-art deep matting methods~\cite{hou2019context,li2020natural,lu2019indices,dai2021learning}, context and dynamic networks are two vital and correlated components. The context includes both global context and local context. Global context intuitively benefits better recognition of the foreground object. It motivates studies on extra context learning modules~\cite{hou2019context,li2020natural}. It is also one of the reasons behind using ASPP or PPM module in recent methods~\cite{lu2019indices,forte2020f,dai2021learning,sun2021semantic}. Local context instead promotes detail capture by caring about correlations within a local region. The convolution operations or dynamic kernels learned from local regions~\cite{lu2019indices,dai2021learning} model local context into the network. On another side, dynamic networks were introduced to matting~\cite{lu2019indices,dai2021learning,li2020natural} to enlarge the model capacity. They also benefit the network in combination with the context assembling~\cite{lu2019indices}.

Since we aim for a more robust matting method, which needs multilevel context information as well as ample model capacity, the first step is taking both context and dynamic networks into consideration efficiently. We show that it is achievable by combining transformer blocks and convolution layers. We investigate %
various 
designs and also provide our insights into them. Considering the lack of training data and a relative large capacity of our model, we study strong data augmentation strategies to prevent overfitting the training data and also generalize the model better.

\textbf{Domain Generalization.} Domain generalization aims at learning better representations that can be transferred to unseen domains. There are many potential solutions, such as data augmentation~\cite{zhong2020random,xu2020robust,zhang2017mixup}, meta learning~\cite{finn2017model,li2018learning}, and adversarial training~\cite{ganin2016domain,luo2019taking}. In deep matting, since only synthetic training data is available, the trained models usually suffer 
from poor
generalization.
They may work well on specific domains, %
such as 
the synthetic ones similar to the training set, but show obvious decreasing performance when applied to another domain, as the examples in Fig.~\ref{fig:example}. Extra data augmentation~\cite{hou2019context,yu2021mask} has been applied to adapt models to real-world images, but they consider limited cases only, such as the resolution gap between the foreground object and background, which may bias the models to those images. We may observe it from Table~\ref{tab:test_benchmark}.

Therefore, we move a step further by rethinking strong data augmentation for matting. We first analyze why current extra augmentation deteriorates the benchmark performance, then propose strong augmentation strategies targeting matting. Our goal is to prevent the model overfitting the synthetic training data and help them generalize better to real-world images.
\section{A Strong Matting Framework with Context Assembling}
\label{sec: transform}

\begin{figure}
    \centering
    \includegraphics[width=\linewidth]{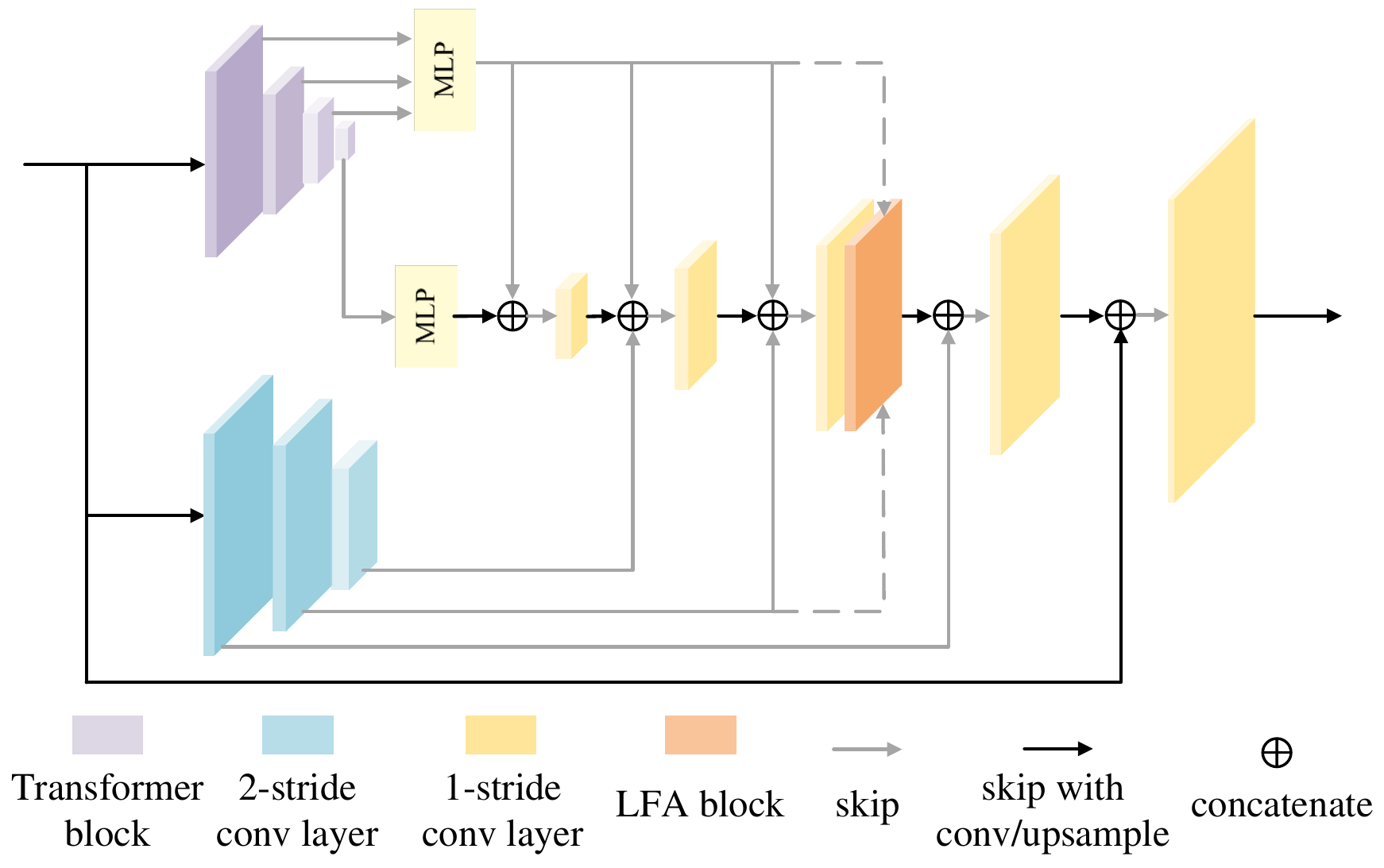}
    \caption{Model architecture of our framework.}
    \label{fig:framework}
    \vspace{-10pt}
\end{figure}

As noted in conventional sampling-based matting~\cite{wang2007optimized,shahrian2013improving} and propagation-based matting~\cite{levin2007closed,chen2013knn}, both nearby and long-distance pixels contribute to alpha prediction depending on their correlations. In deep models, the correlations are related to context. Existing deep matting methods attempt to model contextual attention~\cite{li2020natural} or extra context information~\cite{hou2019context} in the network, while the global context is still under explored. This may limit their performance on complex images such as Fig.~\ref{fig:real_images}. In order to assemble multilevel context information, including global context, we build a baseline combining transformer blocks and convolution layers. Designs of the framework are detailed below:

\textbf{Encoder Design.}
As shown in Fig.~\ref{fig:framework}, the encoder has two branches: a transformer-based branch modeling global context and a convolution-based branch supplementing low-level information for details. 
Driven by recent vision transformers~\cite{dosovitskiy2020image,zheng2021rethinking,wang2021pyramid,xie2021segformer}, we use a $32$-stride pyramid vision transformer backbone to obtain hierarchical features. Since matting models do inference with original image resolutions, fixed position embedding is not suitable for the application. We therefore take advantages of~\cite{xie2021segformer}, where fixed position embedding is replaced with overlapped convolutions. This promises the various input resolutions in matting. 
Due to the large capacity of the transformer blocks, only $2$-stride convolution layers are used in the convolution-based branch to form $8$-stride. We use two small backbones in~\cite{xie2021segformer} (mit-b1 and mit-b2) because of the limited training data for matting. Finally, two encoder architectures with different capacities (E1, E2) are built.

\textbf{Decoder Design.} 
Various decoder designs~\cite{cai2019disentangled,dai2021towards,yu2021mask,li2020natural,lu2019indices,dai2021learning} have been studied in matting models. As the bridge to recover resolutions and capture details, the decoder matters for matting. For instance, previous methods applied feature skip~\cite{lu2019indices}, attention-guided refinement~\cite{li2020natural} or dynamic upsampling~\cite{dai2021learning} to build functional matting decoders aiming at richer details. As the first matting method applying transformers, and considering the importance of the decoder, we investigate an efficient decoder design for our framework.

In general, options for a decoder, in order of decreasing receptive field size, include transformer layers, convolution layers, and MLP layers. Since the transformer branch in the encoder promises a large capacity and global reception field, and to reduce computation as well, we only consider using MLP layers and convolution layers in the basic decoder. These also work well to combining multilevel context information. As a result, several baseline models with different decoders are investigated as listed in Table~\ref{tab:architecture}.

\textbf{Feature Skip Design.} 
Skip information from encoder to decoder has been widely adopted in deep matting methods~\cite{lu2019indices,li2020natural,forte2020f,sun2021semantic}. We observe its importance in our framework as well, and  categorize the skip information into two sources: \textbf{1)} the transformer branch of the encoder (TSkip), where feature maps with different resolutions are skipped to the decoder after MLP/convolution layers. These feature maps transport abundant global information while recovering the resolution. Since the transformer branch starts from $\frac{1}{4}$ resolution, some details may be missing at the initial downsampling stage, so we use \textbf{2)} another source of skip information learned in the convolution branch (LSkip).

\textbf{Low-Level Feature Assembling Attention Block (LFA) Design.} 
Inspired by~\cite{li2020natural,dai2021learning}, where low-level feature maps assist on refining decoder features, we explore efficient low-level feature assembling using a transformer block. It can be naturally extended from the transformer block in the encoder without exhausting designs. Let $Attn(Q,K,V)$ denotes the self-attention operation in the transformer block, the feature fusion attention then can be represent by $Attn(f_{low}, f_{low}, f_{d})$, where $f_{low}$ is the skipped feature from the encoder and $f_{d}$ is the feature in the decoder to be refined. Only one LFA block is added after the $\frac{1}{4}$ resolution decoder layer as noted in Fig.~\ref{fig:framework} in our experiments to restrict the computation. We observe introducing second-order information from encoder to decoder further improves the accuracy.

\section{Domain Generalization and Data Augmentation for Image Matting}
\label{sec: da}
As training images for matting are created using composition, it inevitably results in generalization problems on real-world images. Also, it is noticed that large-capacity transformer-based models may encounter the overfitting problem~\cite{wu2021fastformer,d2021convit}, especially when the dataset is small. Matting datasets~\cite{xu2017deep,qiao2020attention,sun2021semantic}, unfortunately, have limited sizes. They usually use only hundreds of foreground images to generate tens of thousands of synthetic training data, so overfitting is a potential problem. To handle this issue, we study strong augmentation ($\mathcal{SA}$) for better generalization.

Targeting the domain gap between synthetic data and real-world images, extra data augmentation ($\mathcal{DA}$) was proposed~\cite{hou2019context,yu2021mask}. It mainly includes Re-JEPG and Gaussian blur. Experiments in~\cite{hou2019context,yu2021mask} show $\mathcal{DA}$ improves results on real-world images, but deteriorates the performance on the benchmark significantly, as shown in Table~\ref{tab:adobe_benchmark} (CA vs. CA+$\mathcal{DA}$). Therefore, $\mathcal{SA}$ firstly needs to overcome the performance degradation on the benchmark.

\subsection{Rethinking Domain Generalization and Gaps}
\begin{figure}
    \centering
    \mbox
    {
        \subfloat[][]
        {
        \includegraphics[width=0.45\linewidth]{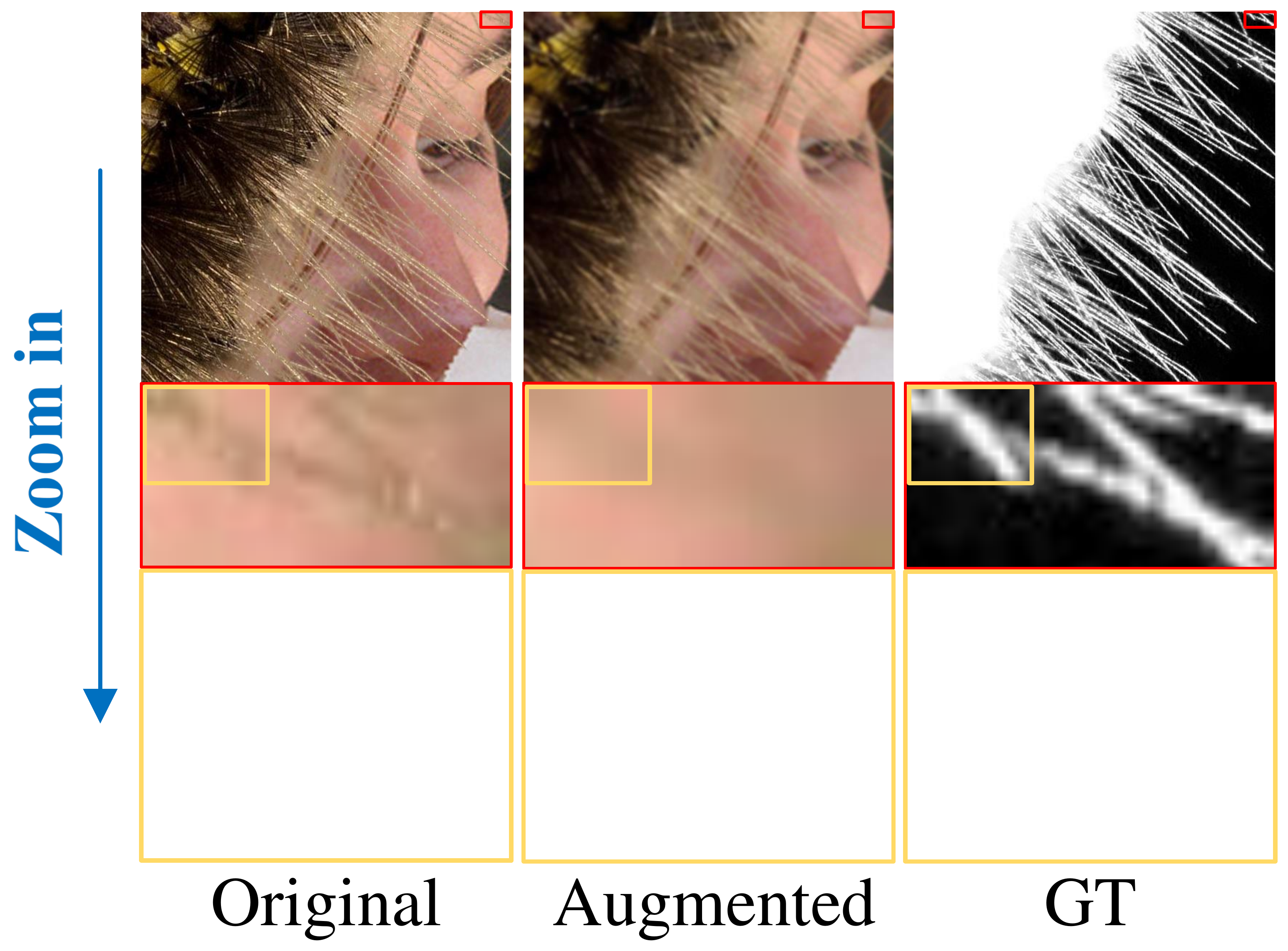}
        \label{fig:showcase}
        }
        \hspace{0.5em}
        \subfloat[][]
        {
         \includegraphics[width=0.45\linewidth]{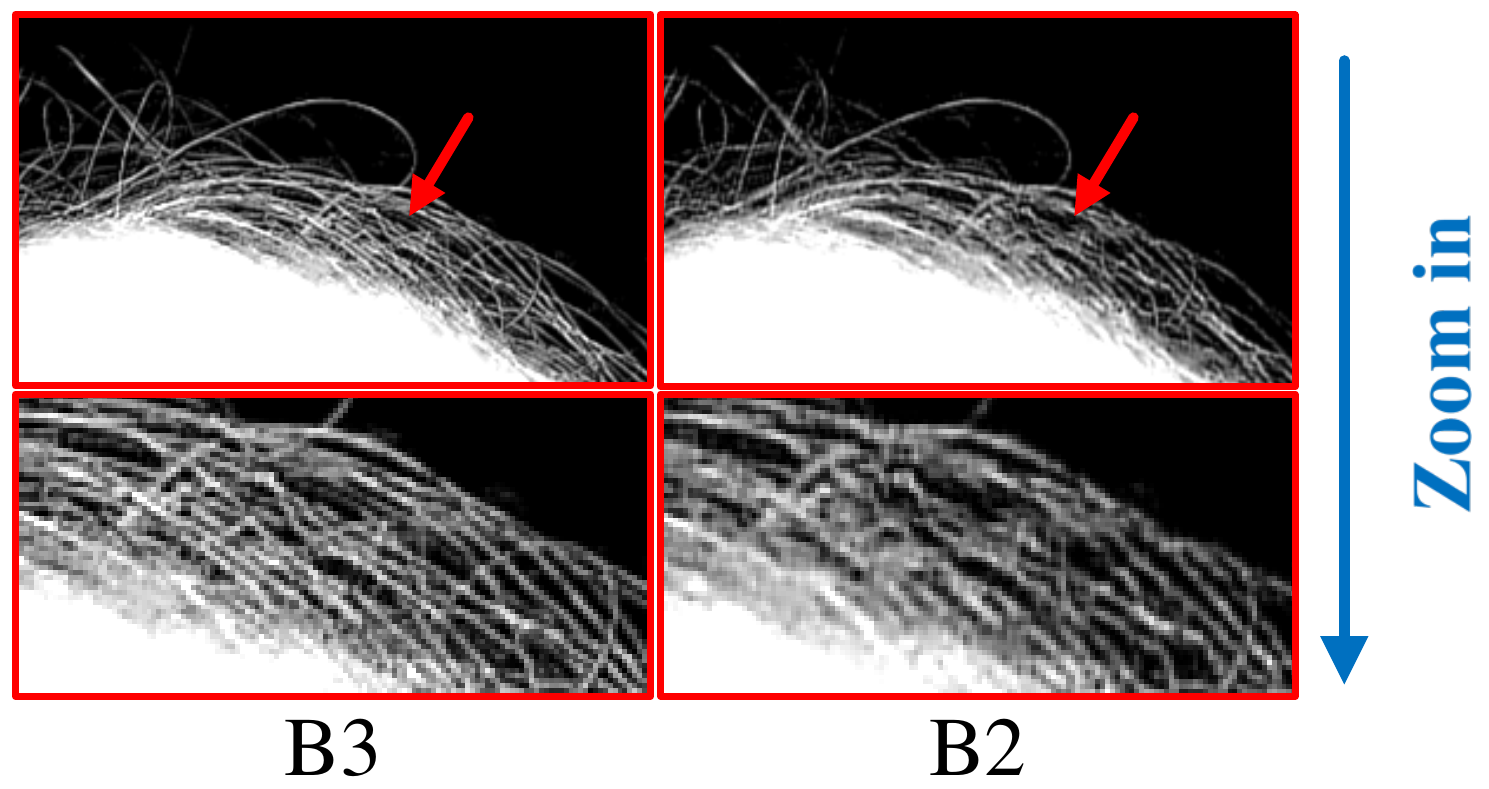}
        \label{fig:showcase_hair}
        }  
    }
    \vspace{-10pt}
    \caption{(a) A comparison between images before and after $\mathcal{DA}$ (Re-JPEG, Gaussian Blur). The augmented image loses its structure and does not match the ground truth. (b) A comparison between B3 and B2 (Table~\ref{tab:architecture}). The hairs in B2 are blurred. }
    \vspace{-20pt}
\end{figure}

\textbf{Why Current Extra Data Augmentation ($\mathcal{DA}$) %
Deteriorates Performance on Benchmarks.
}
An example of using Re-JPEG and Gaussian blur is shown in Fig.~\ref{fig:showcase}. As observed, some background pixels mix values with foreground pixels after augmentation. Their alpha values therefore change from $0$ to a value in range $(0, 1)$. 
This kind of alpha value blending also exists in transparent regions and in some foreground pixels close to the background. In previous works~\cite{hou2019context,yu2021mask}, however, the same alpha ground truths are used after $\mathcal{DA}$, which violates the matting equation. The image after augmentation loses the structure and does not match the alpha ground truth any more. Using these image-alpha pairs for training could mislead the network to wrong predictions. Hence, we argue it is at least one of the main reasons behind the performance deterioration.
To verify this assumption, we carry out a toy-level experiment on DIM. By using Gaussian blur and Re-JPEG with the possibility 0.25 for each as $\mathcal{DA}$, two models are trained: 1) a model trained with $\mathcal{DA}$ using original alpha ground truths; 2) a model trained with $\mathcal{DA}$ using modified alpha ground truths generated by applying the same augmentation as was applied to the RGB image. To ease the difficulty, only L1 alpha prediction loss is applied and fewer training iterations are used. Other training details match the main experiments, as detailed in Section.~\ref{sec: exp}. As present in Table~\ref{tab:toy_rwa}, $\mathcal{DA}$ makes the errors higher, and adjusting the ground truth slightly relieves the problem. Hence, it is at least reasonable to claim that modification of ground truth matters for using $\mathcal{DA}$. The correct ground truth, however, is hard to obtain.


\textbf{What are the Domain Gaps for Matting.}
During the data loading stage, we assume the composition process satisfies the linear Equation~\eqref{matting_eq}. As a result, no matter how the foreground and the background have been processed individually, the image still satisfies the linear equation after the composition. 
Under this assumption, what are the main domain gaps for matting? 

\textbf{1)} \textit{Complexity of Surrounding Context}. For example, the third example in Fig.~\ref{fig:real_images} is a challenging because it is rarely found in the synthetic datasets. 
\textbf{2)} \textit{Source of Images}. In real-world images, foreground and background are from the same source, while this condition is not met in synthetic data. There could be many differences between them: brightness, saturation, sharpness, noise level, etc.
\textbf{3)} \textit{Manual Operations During Photography and Modifications Made to the Images}. For instance, unfocused boundaries, blurred regions,
mosaic generated by image compression, etc. They rarely exist in synthetic datasets.

In this work, we rely on the network to deal with the context domain gap by assembling multilevel context information. As for remaining feature-level gaps, we investigate simple but efficient strong augmentation strategies to generalize the algorithm to real-world images better.

\begin{table}[]\small
    \centering 
    \begin{tabular}[width=\linewidth]{c c | c  c}
    \hline
       $\mathcal{DA}$ & Modified $\alpha$ & SAD  & Grad  \\
       \hline
        & & 32.65 & 18.13  \\ 
        \checkmark &  & 35.79 &  20.16  \\
        \checkmark & \checkmark &  34.00 & 20.12  \\
        \hline
    \end{tabular}
    \caption{$\mathcal{DA}$ using different ground truths. This toy experiment is trained with a batch size of 32, 45k iterations.}
    \vspace{-20pt}
    \label{tab:toy_rwa}
\end{table}

\begin{table*}[]\small
    \centering
    \begin{tabular}{c c c c c | c | c c c c }
    \hline
    No. & Encoder & Decoder & TSkip & LSkip & \#Params  & SAD($\downarrow$)  & MSE($\downarrow$) & Grad($\downarrow$) & Conn($\downarrow$) \\
     \hline
     B1  & E1 & MLP & MLP &  & 13.6M & 39.55  & 0.0102  & 24.22 & 37.35  \\
    B2  & E1 & Conv & MLP &  & 15.4M & 29.85 & 0.0063 & 13.04 & 25.61  \\
   
  \rowcolor{GRAY}
   B3 & E1 & Conv & MLP & \checkmark & 15.8M & 28.94 & 0.0055 & 12.75 & 24.66   \\
    B4 & E1 & Conv & MLPConv & \checkmark  & 18.2M & 29.99 & 0.0064 & 15.98 & 25.86 \\
     B5 & E1 & MLPDW & MLP & \checkmark & 14.0M & 29.79 & 0.0062 & 14.18 & 25.78 \\
   B6  & E1 & MLPDW & MLPConv & \checkmark & 16.1M & 31.45 & 0.0065 & 15.59 & 27.65 \\
    \hline
     
   
    \rowcolor{GRAY}
    B7  & E2 & Conv & MLP & \checkmark & 26.8M & 26.11 & 0.0048 & 10.59 & 21.38  \\
   B8  & E2 & Conv & MLPConv  & \checkmark & 29.2M & 25.66 & 0.0045 & 10.40 & 20.90  \\
    B9 & E2 & MLPDW & MLP & \checkmark & 25.1M & 28.42  & 0.0055 & 12.98  & 24.15 \\
   B10  & E2 & MLPDW & MLPConv & \checkmark & 27.2M & 30.66 & 0.0064 & 14.26 & 26.88 \\
     \hline

    \end{tabular}
    \caption{Ablation study on decoder, feature skip designs on the Composition-1k test set. `MLPDW' denotes `MLP+DepthWise Conv'.
    }
    \vspace{-15pt}
    \label{tab:architecture}
\end{table*}

\begin{table}[]\scriptsize
\small 
    \centering
    \begin{tabular}{c c c c c | c | c  c}
    \hline
     No. & LFA  & $l_{lap}$ & $l_g$ & $l_{gp}$ & \#Params & SAD  & Grad \\
       \hline
      - &  & &  & & 15.8M & 28.94 &  12.75  \\
       - &    & \checkmark & & & 15.8M & 27.64  & 10.68  \\
       - &   &  & \checkmark & & 15.8M & 27.71  & 10.23 \\
       - &   & & & \checkmark & 15.8M & 27.00  & 9.50 \\
       \rowcolor{GRAY}
       N3 & & \checkmark & & \checkmark & 15.8M & 25.86  & 9.69 \\
      -  & \checkmark &  &  & & 16.8M & 27.67  & 12.44 \\
    \rowcolor{GRAY}
      M3  & \checkmark & \checkmark & & \checkmark  & 16.8M & 25.70 & 9.50 \\
        \hline
       - &   & &  & & 26.8M & 26.11 & 10.59 \\
       \rowcolor{GRAY}
      M7 & \checkmark  & \checkmark & & \checkmark & 27.9M  & 25.00 & 9.02  \\
    \hline
    \end{tabular}
    \caption{Ablation study on the LFA module and loss functions on the Composition-1k test set. The upper part(containing \textbf{N3/M3}) and the lower part(containing \textbf{M7}) are based on \textbf{B3} and \textbf{B7}, respectively.}
    \vspace{-25pt}
    \label{tab:extention}
\end{table}

\subsection{Strong Data Augmentation for Matting}
Driven by above analysis, we study $\mathcal{SA}$ for matting. The augmentations are divided into three categories:

1) \textbf{\textit{Linear Pixel-Wise Augmentation.}} By pixel-wise, we mean no interpolation happens on the image. Linear denotes the operations that can be linearly represented. It includes linear contrast, brightness adjustment, noise, etc. Only pixel-level changes happen without any information exchange among different pixels and even different channels. If we look at one channel of location $i$ on image $I$, it can be formulated by:
\begin{equation}\label{pixel-wise}
\begin{aligned}
    I{'}_{i}&=aI_{i}+b \\
&=a\left[\alpha F_{i}+\left( 1-\alpha \right)B_{i}  \right]+\left[\alpha b+\left( 1-\alpha \right)b \right] \\
&=\alpha\left( aF_{i}+b \right)+\left( 1-\alpha \right)\left( aB_{i}+b \right)
\end{aligned}\,,
\end{equation}
where $a$ and $b$ are constant parameters for the linear transformation. According to this equation, linear pixel-wise augmentation obeys Equation~\eqref{matting_eq} no matter it happens on $I_i$, $F_i$ or $B_i$. Augmentation on an image can also be viewed as processing the foreground and background individually. It is natural to extend this equation by:
\begin{equation}\label{pixel-wise-v2}
    I{'}_{i}=\alpha\left( aF_{i}+b \right)+\left( 1-\alpha \right)\left( mB_{i}+n \right) \,,
\end{equation}
where different linear transformations happen on the foreground and the background.

2) \textbf{\textit{Nonlinear Pixel-Wise Augmentation.}} In the opposite to linear operations, there are also non-linear augmentations, such as gamma correction, hue/saturation adjustment, etc. Due to their nonlinear nature, Equation~\eqref{matting_eq} is violated if the augmentations happen on $I$.

3) \textbf{\textit{Region-Wise Augmentation.}} Region-Wise augmentation means operations applied using multiple pixels. For instance, blur, jpeg compression, etc. After interpolations on $I$, Equation~\eqref{matting_eq} is violated, which needs alpha ground truth to be modified accordingly.

Based on this categorization, we propose strong data augmentation strategies:

i) \textbf{{Augment the Foreground Alone (AF).}} Motivated by the random jitter in~\cite{li2020natural}, augmenting foreground alone is effective and obeys the composition equation. The ground truth does not need to be modified no matter which augmentation is taken because it happens before composition.

ii) \textbf{{Augment the Foreground and the Background Individually (AFB).}} It is an extended version of option i) and inspired by Equation~\eqref{pixel-wise-v2}. Through augmenting foreground and background individually before composition, the linear composition equation is still satisfied, the ground truth alpha matte hence does not need to be modified no matter which augmentation is taken.

iii) \textbf{{Augment the Composited Image (AC).}} This strategy can be further divided into two sub types. If linear pixel-wise augmentation is applied, the composition equation is satisfied as Equation~\eqref{pixel-wise-v2}. Using other strategies instead violates the equation, where a new ground truth is needed. Due to the expense of obtaining the real ground truth, we propose to generate pseudo label to facilitate the training. The strategy is to predict the pseudo label using the parameters from the last training iteration by rotating or channel-shuffling the input to generate a new training sample. We anticipate this operation promotes the network to learn features of the augmented images without sacrificing accuracy.

\section{Experiments and Discussions}
\label{sec: exp}

\subsection{Implementation Details}
Our models are trained on the deep image matting (DIM) dataset~\cite{xu2017deep} only. It contains $431$ foreground images in the training set and $50$ foreground images in the test set. We generate the training samples using background images randomly selected from MS COCO~\cite{lin2014microsoft}, and use the same rules as~\cite{xu2017deep} to produce test images using background images selected from Pascal VOC~\cite{everingham2010pascal}. The evaluation metrics are commonly-used Sum of Absolute Differences (SAD), Mean Squared Error (MSE), Gradient (Grad) error, and Connectivity (Conn) error. Implementation of~\cite{xu2017deep} is used.

\textbf{Training Details.}
Our baseline models follow the dataloader pipeline in~\cite{li2020natural}. To be specific, the 4-channel input concatenates the RGB image and the trimap. The RGB image is generated on-the-fly through the following basic augmentation: foreground random affine, foreground random combination, random resize, random crop, foreground random jitter, and composition. More details are explained in~\cite{li2020natural}. $512\times512$ patches are finally generated for training. We initialize the weights of the mit backbones using the pretrained weights on ImageNet-1K\cite{deng2009imagenet} from~\cite{xie2021segformer} for the transformer branch. Other parameters are initialized with Xavier. The training stage is optimized by AdamW~\cite{loshchilov2017decoupled} optimizer using initial learning rate $6\times10^{-4}$ with cosine decay. The warm up stage takes $1000$ iterations. Without specially clarifying, we update parameters for $90k$ iterations with a batch size of $32$. Batch size $64$ and $120k$ iterations are used for final benchmark results, as detailed in Table~\ref{tab:adobe_benchmark} and~\ref{tab:test_benchmark}. 

\textbf{Loss Functions.}
Our baseline models only use L1 alpha prediction loss and composition loss as~\cite{lu2019indices}. Since other loss functions, such as laplacian loss ($l_{lap}$) and gradient loss ($l_g$), are applied in previous pure convolution-based methods~\cite{forte2020f, hou2019context}, here we validate their effects in our framework. Besides using the usual $l_{lap}$ and $l_g$, we define a new gradient loss with gradient penalty($l_{gp}$) for local smoothness:
\begin{equation}
\begin{aligned}
    l_{gp}&=\left\| \nabla \alpha_{x}-\nabla \hat{\alpha}_{x} \right\|_{1}+\left\| \nabla \alpha_{y}-\nabla \hat{\alpha}_{y} \right\|_{1} \\
    &+\lambda\left(  \left\| \nabla \alpha_{x} \right\|_{1}+\left\| \nabla \alpha_{y} \right\|_{1}\right)
\end{aligned} \,,
\end{equation}
where $\lambda$ is set as $0.01$ in our experiments. 

\begin{table}[] \small
    \centering
    \addtolength{\tabcolsep}{-2pt}
    \begin{tabular}{l c c c c r}
    \hline
      Method  & SAD & MSE & Grad & Conn &  \# Params \\
      \hline
      CF~\cite{levin2007closed} & 168.1 & 0.091 & 126.9 & 167.9 & - \\
      KNN~\cite{chen2013knn} & 175.4 & 0.103 & 124.1 & 176.4 &  - \\
       DIM~\cite{xu2017deep} & 50.4 & 0.014 & 31.0 & 50.8 & $>130.55$M  \\ 
       IndexNet~\cite{lu2019indices} & 45.8 & 0.013 & 25.9 & 43.7  &  8.15M \\
       CA~\cite{hou2019context} & 35.8 & 0.0082 & 17.3 & 33.2 & 107.5M \\
       CA+$\mathcal{DA}$~\cite{hou2019context} & 71.3 & 0.0236 & 38.8 & 72.0 & 107.5M \\
       GCA~\cite{li2020natural} & 35.28 & 0.0091 & 16.9 & 32.5 & 25.27M \\
       
        A$^2$U~\cite{dai2021learning} & 32.15 & 0.0082 & 16.39 & 29.25 & 8.09M \\
        SIM~\cite{sun2021semantic} & 28.0 & 0.0058 & 10.8 & 24.8 & 70.16M \\ 
        FBA~\cite{forte2020f} & 26.4 & 0.0054 & 10.6 & 21.5 & 34.69M \\
        FBA+TTA~\cite{forte2020f} & 25.8 & 0.0052 & 10.6 & 20.8 & 34.69M \\
        \hline
        M3\textsuperscript{$\ddagger$} & 23.98 & 0.0042 & 8.54 & 18.88 & 16.8M \\
        M7\textsuperscript{$\ddagger$} & \textbf{22.87} & \textbf{0.0039} & \textbf{7.74} & \textbf{17.84} & 27.9M \\
       \hline
    \end{tabular}
    \vspace{2pt}
    \caption{Benchmark results on the Composition-1k test set. The best performance is in boldface. $\ddagger$ denotes training with a batch size of 64, 120k iterations using our $\mathcal{SA}$.}
    \vspace{-20pt}
    \label{tab:adobe_benchmark}
\end{table}

\subsection{Results on the Deep Image Matting Dataset}

\textbf{Ablation Study on Model Architecture.} 
Based on the two-branch encoder, here we investigate designs of the decoder, the skip layer and the additional attention module. 

According to the results in Table~\ref{tab:architecture} and Table~\ref{tab:extention}, we draw the following observations: 
\textbf{1)} Compared with the MLP layer and the MLPDW layer, the Conv layer suits the decoder of matting better (B1 vs. B2, B5 vs.B3, B6 vs. B4, B9 vs.B7, B10 vs. B8); 
\textbf{2)} Skipped information from the transformer branch to the decoder can be efficiently achieved by a simple MLP layer (B3 vs. B4, B5 vs. B6, B9 vs. B10); 
\textbf{3)} Low-level skip fusion is important for recovering details (B2 vs. B3, also see Fig.~\ref{fig:showcase_hair});
\textbf{4)} Additional low-level feature assembling attention module further improves the results (Table~\ref{tab:extention}); 
\textbf{5)} The advantage of larger backbone is gradually weakened with improvement on the architecture and loss functions (B3$\xrightarrow{}$M3 vs. B7$\xrightarrow{}$M7 in Table~\ref{tab:extention}).

\textbf{Ablation Study on Loss Functions.}
Here we justify effectiveness of laplacian loss ($l_{lap}$), gradient loss ($l_{g}$) and the proposed gradient loss with gradient penalty ($l_{gp}$) in our framework. Results are reported in Table~\ref{tab:extention}. Compared with using only basic loss functions, $l_{lap}$, $l_{g}$, and $l_{gp}$ all reduce the errors, and our $l_{gp}$ works better than normal $l_{g}$. Combining $l_{lap}$ and $l_{gp}$ together builds the best prediction. We use these two losses in the following experiments.

\textbf{Comparison with State of the Art.}
Benchmark results on the Composition-1k are list in Table~\ref{tab:adobe_benchmark}. Our models achieve significantly better results on all the metrics. Compared with a currently top-performing FBA+TTA~\cite{forte2020f} model, our method (M7\textsuperscript{$\ddagger$}) gains $11\%$ improvement on SAD and $27\%$ improvement on Grad without any augmentation on the test images. Moreover, our method is more robust to trimap precision, as shown in Fig.~\ref{fig:real_images} and ~\ref{fig:trimap_sad}, and the detailed evaluations in the supplement.

\subsection{Generalization on Various Benchmarks}
To verify the generalization ability of matting methods to unseen domains, comparison experiments are carried out in Table~\ref{tab:test_benchmark}. Specifically, we test models trained merely with the DIM dataset~\cite{xu2017deep} on several different benchmarks without fitting on their training set (except that SIM~\cite{sun2021semantic} is trained on SIMD~\cite{sun2021semantic}, which has $763$ foregrounds, $332$ more foregrounds than DIM). The test benchmarks include Distinction-646~\cite{qiao2020attention}, SIMD~\cite{sun2021semantic}, and AIM-500~\cite{li2021deep}. Distinction-646 and SIMD are synthetic benchmarks, and AIM-500 is a real-world one but with simple scenes, so none of them alone is perfect for measuring generalization ability of a model. However, since they contain images from different sources and various domains, it is at least reasonable to combine them together to see how algorithms perform quantitatively on all of them, and the overall results should reflect how well a model can adapt to diverse images to some extent. Note that, since SIMD has only provided alpha ground truths and foregrounds until the submission, we generate the test set following the rule of~\cite{xu2017deep} and name it as SIMD$_{our}$. To ensure all the methods can be test on a normal modern graphic card, we restrict the maximum length of the test images in SIMD$_{our}$ by $2000$. 

\begin{figure}[!t]
    \centering
    \includegraphics[width=0.65\linewidth]{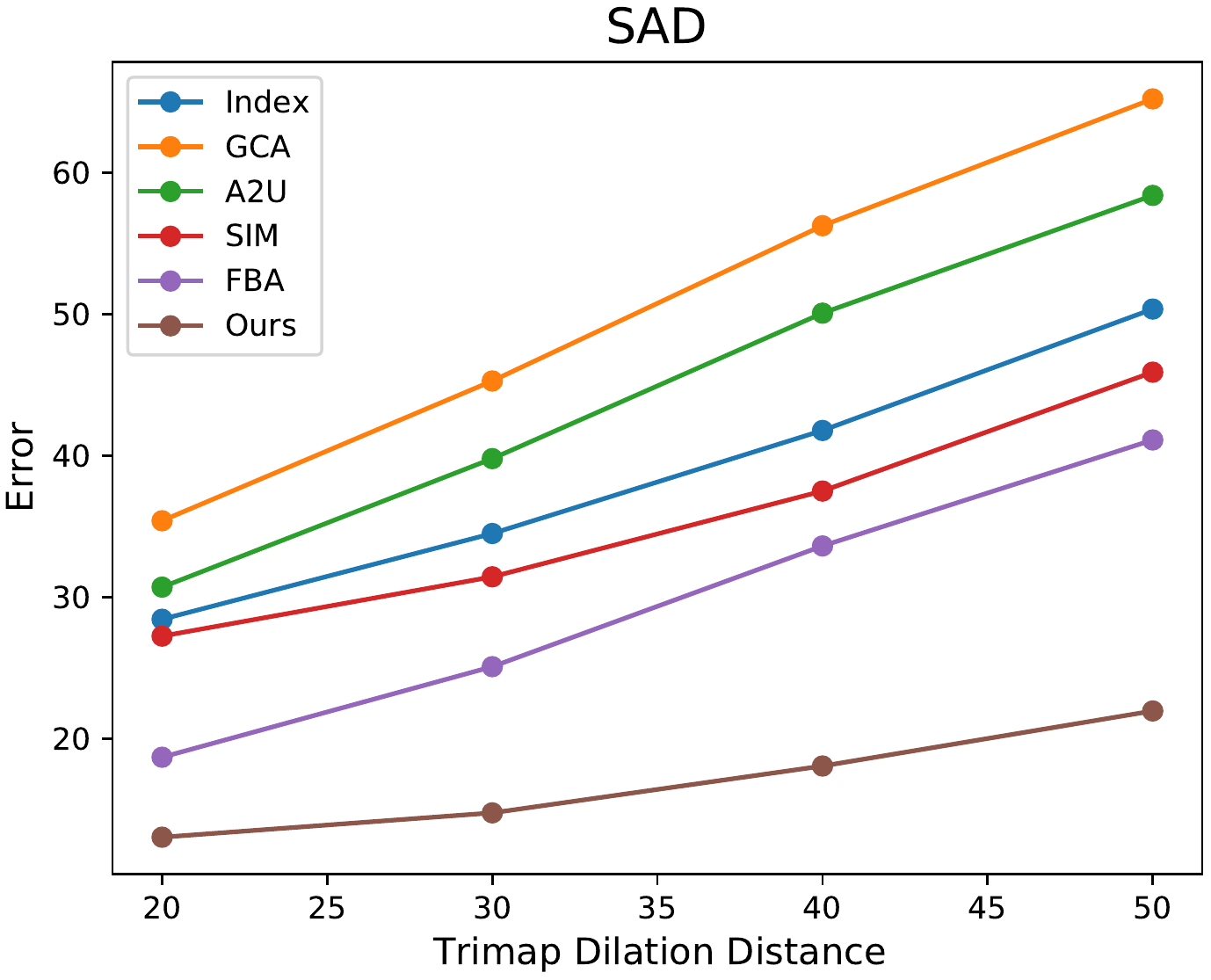}
    \vspace{-5pt}
    \caption{Robustness to trimap precision on AIM-500}
    \label{fig:trimap_sad}
    \vspace{-15pt}
\end{figure}

\textbf{Ablation Study on Strong Data Augmentation.}
Here we investigate the $\mathcal{SA}$ strategies. We either use AF, AFB alone, or combine them with AC. Specifically, the linear pixel-wise augmentations include: \textit{linear contrast, brightness adjustment, channel inversion/shuffling, gaussian/poisson noise, random dropout, cloud, snow, multiply, salt and pepper}; the nonlinear pixel-wise augmentations include: \textit{gamma contrast, hue and saturation add on, histogram equalization}; and region-wise augmentations consist of \textit{gaussian blur and jpeg compression}. If AF or AFB is applied alone, we set the possibility as 0.5 and keep the ground truths unmodified; if they are combined, possibility of each is changed to 0.25; further, if AC is added on, we set its possibility as 0.1 when AF and AFB do not happen, and generate pseudo labels for the augmented samples as explained in Section~\ref{sec: da} when needed. More details are in supplement. As shown in Table~\ref{tab:test_benchmark}, both AF or AFB improve the AIM-500 results, especially AFB, but they also make errors on Distinction-646, SIMD$_{our}$, and Composition-1k (supplement) slightly higher. AF is more stable on synthetic benchmarks compared with AFB. Hence, we carry out `AF+AFB'. It averages the effects of AF and AFB. Based on `AF+AFB', AC further improves results on AIM-500 and keeps results on other synthetic benchmarks comparable, so we use `AF+AFB+AC' as the final $\mathcal{SA}$.

\begin{table*}[]\scriptsize
    \centering
    \small 
    \addtolength{\tabcolsep}{-2pt}
    \begin{tabular}{l c c c | c c c c | c c c c | c c c c}
    \hline
     & & & & \multicolumn{4}{c|}{Distinction-646~\cite{qiao2020attention}} & \multicolumn{4}{c|}{SIMD$_{our}$~\cite{sun2021semantic}} & \multicolumn{4}{c}{AIM-500~\cite{li2021deep}} \\
       \multicolumn{4}{c|}{Method}  & SAD & MSE & Grad & Conn & SAD & MSE & Grad & Conn & SAD & MSE & Grad & Conn \\
      \hline
        \multicolumn{4}{c|}{IndexNet\textsuperscript{*}~\cite{lu2019indices}}  & 42.64 & 0.0256 & 40.17 & 42.76 & 92.45 & 0.0388 & 45.85 & 93.14 & 28.49 & 0.0288 & 18.15 & 27.95 \\
        \multicolumn{4}{c|}{CA\textsuperscript{*}~\cite{hou2019context}}   & 49.07 & 0.0557 & 114.77 & 48.27 & 79.46 & 0.0291 & 51.03 & 77.88 & 26.33 & 0.0266 & 18.89 & 25.05 \\
        \multicolumn{4}{c|}{CA+$\mathcal{DA}$\textsuperscript{*}~\cite{hou2019context}}  & 46.03 & 0.0356 & 55.45 & 46.18 & 102.97 & 0.0469 & 74.39 & 103.52 & 32.15 & 0.0388 & 30.25 & 31.00 \\
        \multicolumn{4}{c|}{GCA\textsuperscript{*}~\cite{li2020natural}}  & 31.00 & 0.0171 & 21.19 & 29.62 & 75.81 & 0.0271 & 40.57 & 74.45 & 35.10 & 0.0389 & 25.67 & 35.48 \\
       
         \multicolumn{4}{c|}{A$^2$U\textsuperscript{*}~\cite{dai2021learning}}  & 28.74  & 0.0143 & 17.42  & 27.62 & 68.70 & 0.0268 & 39.00 & 66.76 & 30.38 & 0.0307 & 22.60 & 30.69 \\
         \multicolumn{4}{c|}{SIM\textsuperscript{*}~\cite{sun2021semantic}}  & \textbf{22.68} & 0.0137 & 20.11 & \underline{21.03} & \textcolor{blue}{\textbf{37.07}}  & \textcolor{blue}{\underline{0.0099}} & \textcolor{blue}{22.29} & \textcolor{blue}{\underline{33.30}}  & 27.05 & 0.0311 & 23.68 & 27.08 \\ 
         \multicolumn{4}{c|}{FBA\textsuperscript{*}~\cite{forte2020f}} & 30.70 & 0.0150 & 18.89 & 29.65 & 41.55 & 0.0109 & 23.21  & 35.07 & 19.05 & 0.0162 & 11.42 & 18.30 \\
      \hline
    
        & AF & AFB & AC &  & & & &  &  &  &  & & & & \\
        
        N3 & & & & 25.86 & 0.0105 & 13.25 & 24.11 & 41.83 & \underline{0.0099} & 22.19 & 34.98 & 16.08 & 0.0122 & 11.69 & 15.55 \\
        N3 & \checkmark & & & 25.18 & 0.0108 & 13.45 & 23.35 & 44.14 & 0.0122 & 23.46 & 37.96 & 16.38 & 0.0112 & 10.75 & 16.00 \\
        N3 &  & \checkmark &  & 27.02 & 0.0123 & 14.44 & 25.34 & 44.46 & 0.0113 & 23.93 & 38.30 & \textbf{13.46} & 0.0097 & 9.98 & \textbf{12.47} \\
         N3 & \checkmark & \checkmark &  & 26.46 & 0.0117 & 14.18 & 24.87 & 42.89 & 0.0110 & 23.37 & 36.05 & 14.63 & 0.0108 & 10.46 & 13.75 \\
         N3 & \checkmark & \checkmark & \checkmark & 26.32 & 0.0119 & 14.40 & 24.56 & 43.49 & 0.0109 & 23.96 & 36.84 & 14.18 & \underline{0.0093} & \underline{9.36} & 13.61 \\
        M3\textsuperscript{$\ddagger$} & \checkmark & \checkmark & \checkmark & 23.67 & \underline{0.0100} & \underline{11.30} & 21.37 & 42.68 & 0.0121 & \underline{21.20} & 35.84 & \underline{13.68} & \textbf{0.0091} & \underline{9.36} & \underline{13.06} \\
        M7\textsuperscript{$\ddagger$} & \checkmark & \checkmark & \checkmark & \underline{23.25} & \textbf{0.0097} & \textbf{11.09} & \textbf{21.00} & \underline{37.31} & \textbf{0.0090} & \textbf{20.00} & \textbf{30.10} & 13.97 & 0.0094 & \textbf{8.89} & 13.21 \\
       \hline
    \end{tabular}
    \vspace{2pt}
    \caption{Generalization results on the Distinction-646, SIMD$_{our}$ and AIM benchmarks. The best performance is in boldface. The second is underlined.
    \textbf{All the models are merely trained with the DIM training set ($431$ foregrounds), except that SIM is trained with the SIMD training set  ($736$ foregrounds), so we set SIM's results on SIMD as \textcolor{blue}{blue color} to represent SIM is trained with this dataset.}
    \textsuperscript{*} means using the officially provided model.
    $\ddagger$ denotes training with a batch size of  64, 120k iterations using our $\mathcal{SA}$ (AF+AFB+AC).}
    \label{tab:test_benchmark}
    \vspace{-20pt}
\end{table*}

\begin{figure}[!t]
    \centering
    \includegraphics[width=\linewidth]{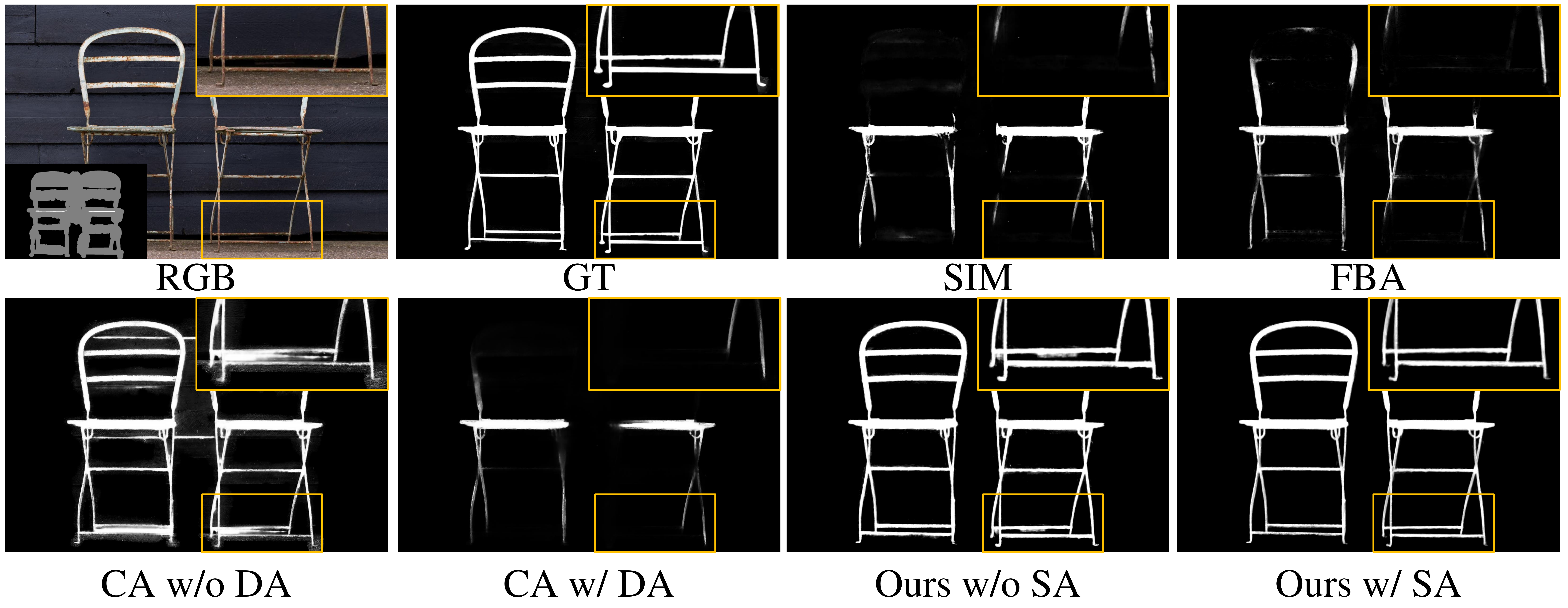}
    \vspace{-10pt}
    \caption{Visual results on the AIM-500 benchmark. The methods in comparison are SIM~\cite{sun2021semantic}, FBA~\cite{forte2020f}, CA w/o $\mathcal{DA}$~\cite{hou2019context}, CA w/     $\mathcal{DA}$~\cite{hou2019context}, Ours w/o $\mathcal{SA}$, Ours w/ $\mathcal{SA}$.}
    \label{fig:aim_1}
    \vspace{-15pt}
\end{figure}

\textbf{Comparison with State of the Art.}
Compared with other methods, our M3\textsuperscript{$\ddagger$} ans M7\textsuperscript{$\ddagger$} models achieve best performance on all three benchmarks in Table~\ref{tab:test_benchmark}, especially with MSE and Grad metrics. Note that, $\mathcal{DA}$ in~\cite{hou2019context} degrades its performance on Composition-1k (Table~\ref{tab:adobe_benchmark}), SIMD$_{our}$ and AIM-500 significantly, even though AIM-500 is a real-world benchmark. Our  $\mathcal{SA}$ instead promises comparable results on the synthetic benchmarks and much better results on the real-world benchmark. The advantages of $\mathcal{SA}$ against $\mathcal{DA}$ can also be noticed from visual examples in Fig.~\ref{fig:example} and~\ref{fig:aim_1}. Moreover, when longer training time is taken, stable improvements are observed. The examples in Fig.~\ref{fig:real_images} and~\ref{fig:aim_1} further verify the effectiveness of our model and $\mathcal{SA}$.
More visual results on real-world images and benchmarks are shown in the supplement.

\subsection{Results on the $\tt alphamatting.com$}
We show the results of M7\textsuperscript{$\ddagger$} on the $\tt al\-ph\-a\-ma\-tting.com$~\cite{rhemann2009perceptually} online benchmark in Table.~\ref{tab:alphamatting}. Note that, SIM is trained with the SIMD training set, which has $736$ foregrounds in the training set, while DIM only has $431$ foregrounds in the training set; GCA and A$^2$U retrain their models with the whole DIM dataset (including both training set and test set). Our result is directly reported from M7\textsuperscript{$\ddagger$} without using extra data or fine-tuning the model, but it still achieves top-performing ranks, especially on MSE and Grad. See the full table in the supplement.

\begin{table}[!h]\tiny
    \centering
    \begin{tabular}{l c c c c c c c c}
    \hline
       Method  & \multicolumn{4}{c}{MSE} & \multicolumn{4}{c}{Grad}  \\
         & overall  & S & L & U  & overall & S & L & U \\
    \hline
    Ours-M7\textsuperscript{$\ddagger$} & \textbf{6.8} &\textbf{5.9}  & \textbf{5.5} & \underline{9.1} & \textbf{4.7} & \textbf{4.8} & \textbf{3.8} & \textbf{5.5} \\
    SIM~\cite{sun2021semantic} & \underline{7} & \underline{8.1} & \textbf{5.5} & \textbf{7.4} & \underline{6.9} & \underline{8.5} & \underline{5.9} & \underline{6.5} \\
    A$^2$U~\cite{dai2021learning} & 15.5 & 13 & 12.6 & 20.8 & 12.3 & 11.3 & 9.4 & 16.1 \\
    GCA~\cite{li2020natural} & 15.3 & 15.1 & 14.5 & 16.4 & 13.7 & 13.6 & 12.5 & 15 \\
    CA~\cite{hou2019context} & 17.6 & 20.9 & 18.6 & 13.3 & 14.6 & 15.8 & 15.5 & 12.6 \\
    IndexNet~\cite{lu2019indices} & 22.9 & 25.3 & 21.5 & 22 & 18.6 & 17.3 & 17.3 & 21.4 \\
    \hline
    \end{tabular}
    \caption{Results on the $\tt alphamatting.com$ online benchmark.}
    \label{tab:alphamatting}
    \vspace{-15pt}
\end{table}

\section{Conclusion}
We propose \ourmethod, a matting method showing higher robustness to various trimap precision and images from different domains. The efforts behind this include a new matting framework and strong augmentation strategies specifically designed for matting. We first build the strong baseline by assembling multilevel context information, then analyse the problems behind current data augmentation and design strong augmentation strategies. To verify generalization capability of the model, we not only show visual results on real-world images, but also design a series of evaluation experiments on several benchmarks without fitting their training sets. Our method achieves state-of-the-art results on all the benchmarks. We hope our work opens up more possibilities for future works on deep matting.

\textbf{Limitations}
There are still many %
challenging 
cases, such as strong light in the background, cannot be handled by our method. We show failure cases in the supplement. To %
tackle 
those cases, we may need to better learn the structure of the foreground objects. We leave it as %
future work.

{\small
\bibliographystyle{ieee_fullname}
\bibliography{egbib}
}


\appendix

\section{Boosting Robustness of Image Matting with Context Assembling and \\
Strong Data Augmentation: 
Supplementary Material}

\section{List of Content}
This supplementary material includes the following contents:
\begin{itemize}
    \item Measure of GFLOPs.
    \item Robustness to trimap precision.
    \item Details on $\mathcal{SA}$ strategies.
    \item Ablation study on $\mathcal{SA}$ on the Composition-1k.
    \item More visual results on real-world images and benchmarks, including Composition-1k~\cite{xu2017deep}, Distinction-646~\cite{qiao2020attention}, SIMD$_{our}$~\cite{sun2021semantic}, AIM-500~\cite{li2021deep}.
    \item Results on the $\tt alphamatting.com$~\cite{rhemann2009perceptually} benchmark.
    \item Failure cases.
\end{itemize}

\section{Measure of GFLOPs}
We measure GFLOPs of SIM~\cite{sun2021semantic}, FBA~\cite{forte2020f} and our method. Results are reported in Table.\ref{tab:flop}.
\begin{table}[!h]\small
    \centering
    \begin{tabular}{l|c}
    \hline
        Method & GFLOPs \\
        \hline
       SIM~\cite{sun2021semantic}  & 48.30 \\
       FBA~\cite{forte2020f} & 30.47 \\
       \hline
       M3\textsuperscript{$\ddagger$} & 16.62 \\
       M7\textsuperscript{$\ddagger$} & 22.22 \\
       \hline
    \end{tabular}
    \caption{GFLOPs measured on a $224\times224$ input.}
    \label{tab:flop}

\end{table}

\section{Robustness to Trimap Precision}
\begin{figure*}
    \centering
    \mbox
    {
        \subfloat[][]
        {
        \includegraphics[width=0.48\linewidth]{latex/figures/SAD_trimap}
        \label{fig:sad_trimap}
        }
        \hspace{0.5em}
        \subfloat[][]
        {
         \includegraphics[width=0.48\linewidth]{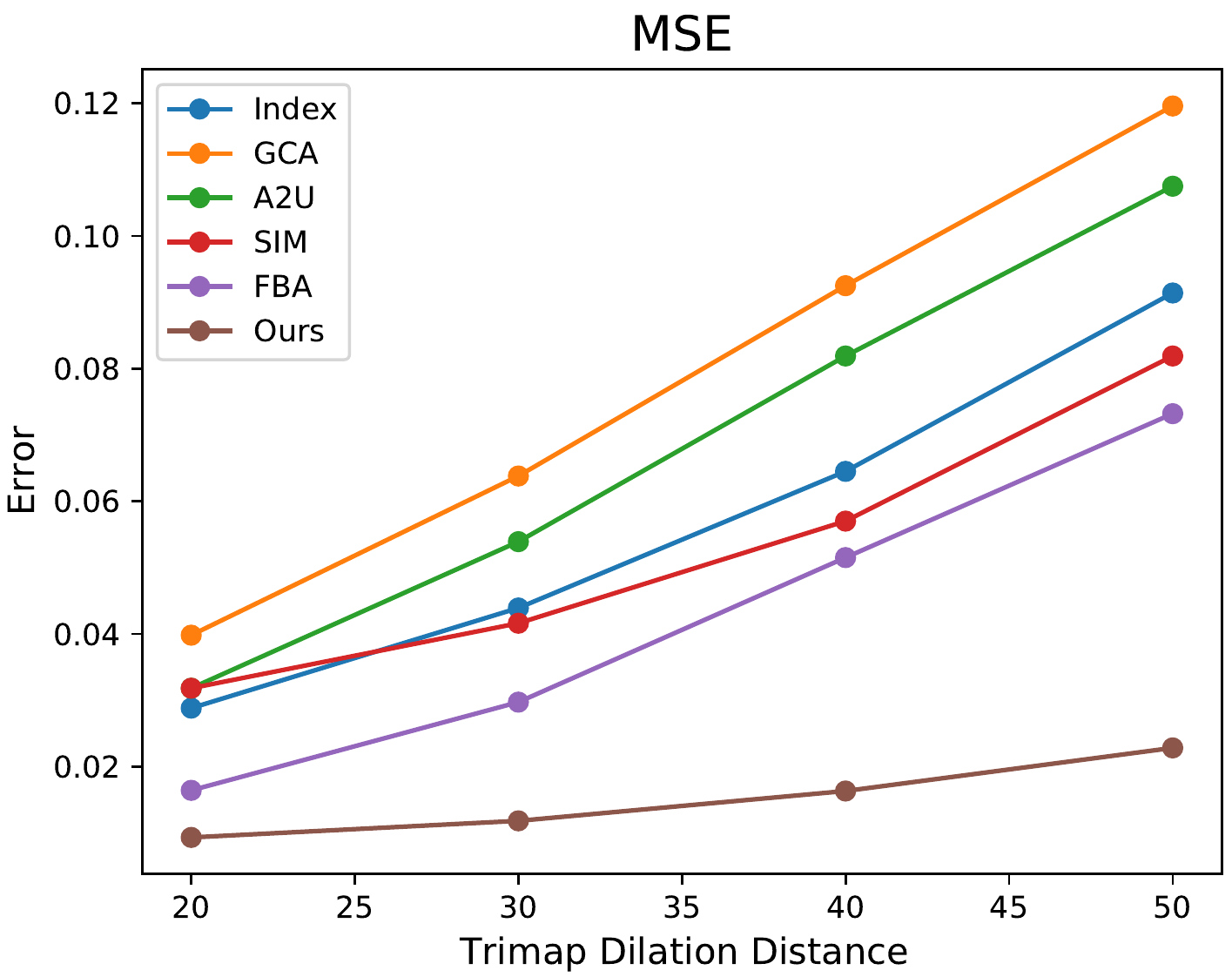}
        \label{fig:mse_trimap}
    }
    }
    \mbox
    {
      
        \subfloat[][]
        {
         \includegraphics[width=0.48\linewidth]{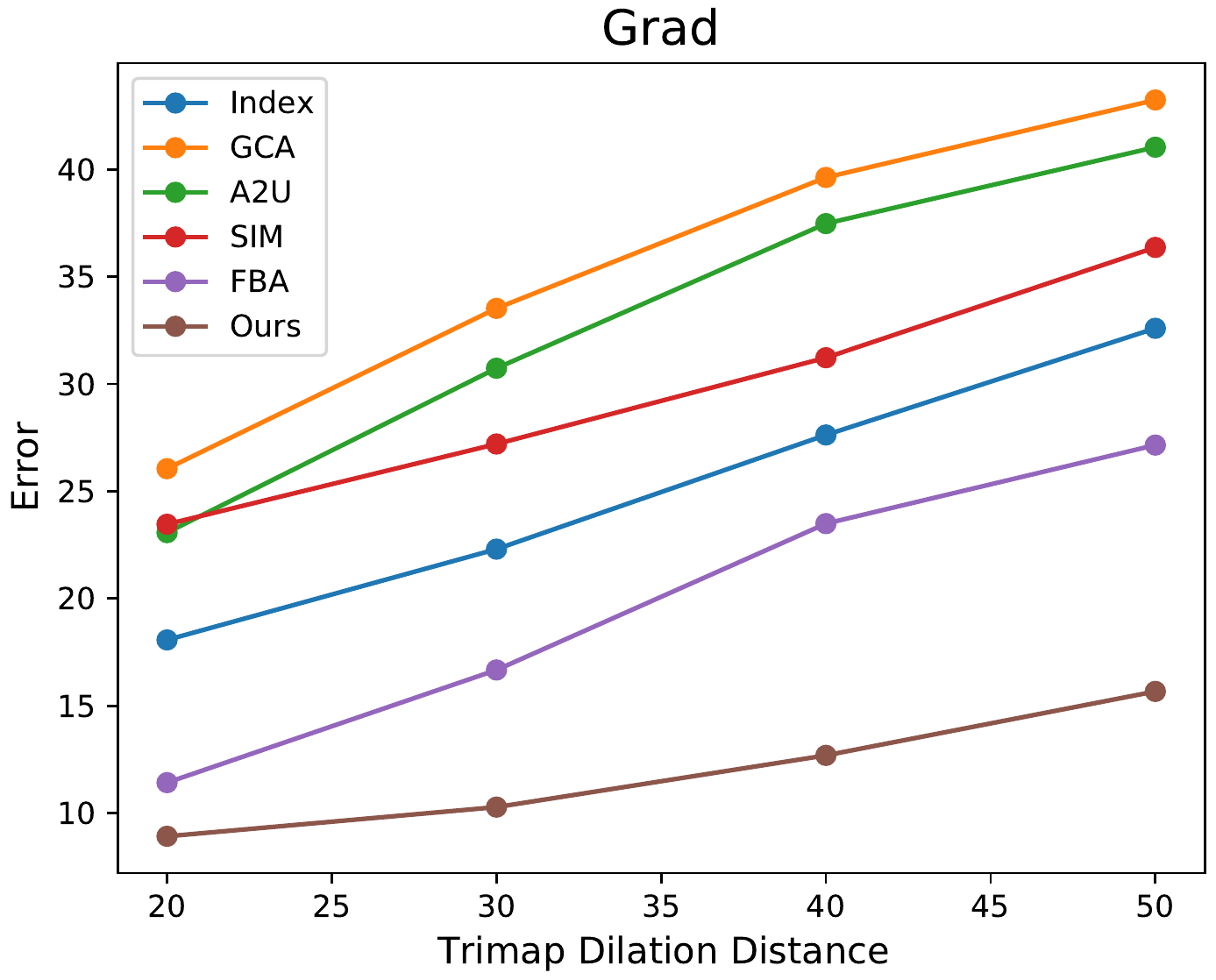}
        \label{fig:grad_trimap}
        } 
        \subfloat[][]
        {
         \includegraphics[width=0.48\linewidth]{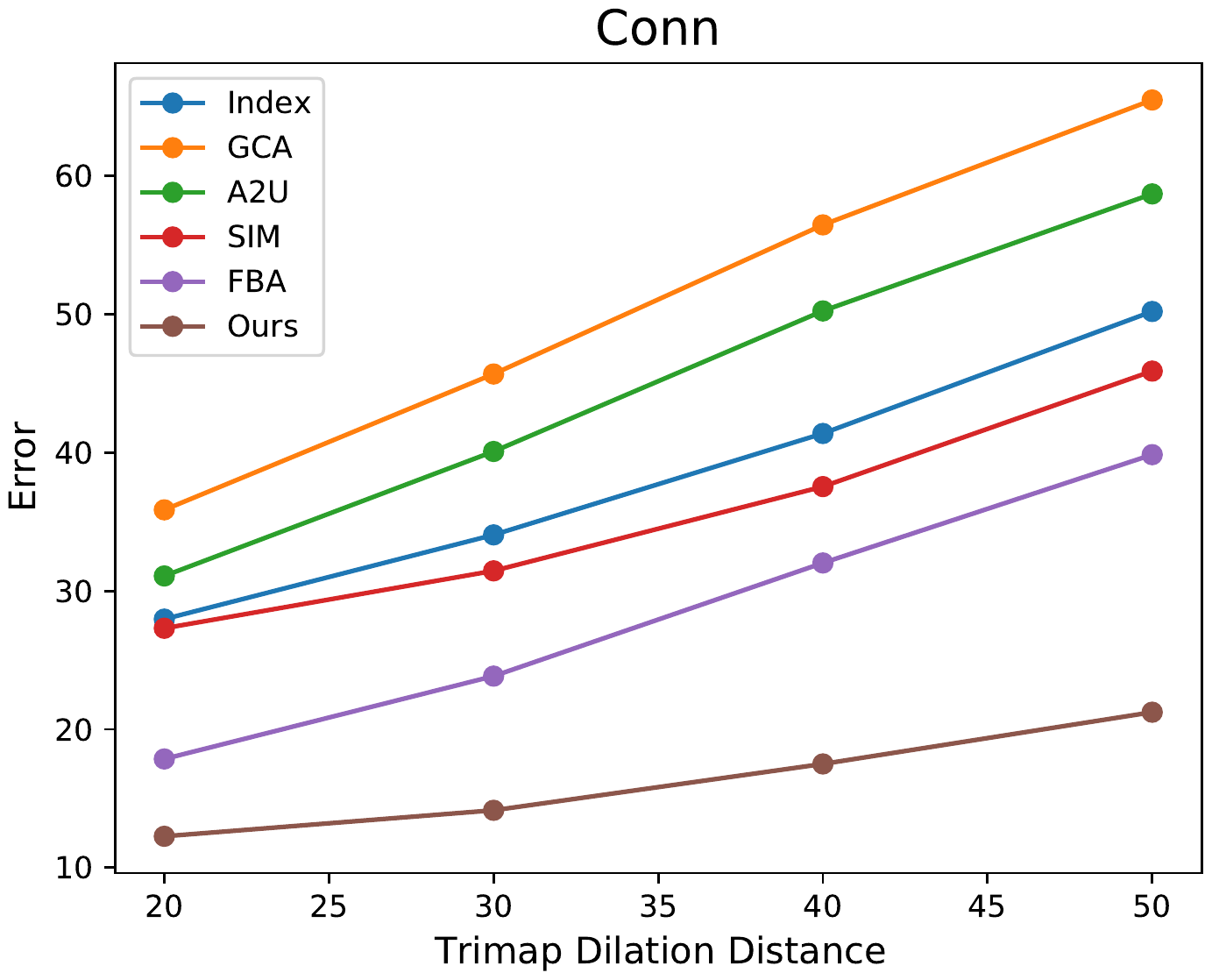}
        \label{fig:conn_trimap}
        } 
    }
    \caption{Robustness to trimap precision on the AIM-500.}
    \label{fig:robustness_trimap}

\end{figure*}

We conduct evaluations on the AIM-500 with different trimap dilation distances. Methods in comparison are IndexNet~\cite{lu2019indices}, GCA~\cite{li2020natural}, A$^2$U~\cite{dai2021learning}, SIM~\cite{sun2021semantic}, FBA~\cite{forte2020f} and our M7\textsuperscript{$\ddagger$}. In detail, we generate $4$ sets of trimaps using random dilation distances within $[11, 20]$, $[21, 30]$, $[31, 40]$, $[41, 50]$, respectively. We denote them as $20$, $30$, $40$, $50$ accordingly in Fig.~\ref{fig:robustness_trimap}. As shown in Fig.~\ref{fig:robustness_trimap}, our method is obviously more robust to varying trimap precision on all the metrics.

\section{Details on the Strong Data Augmentation Strategies}
To supplement the content in the main text, we further detail the $\mathcal{SA}$ strategies in our experiments here.

If AF or AFB is applied alone, we set the possibility as 0.5 and keep the ground truths unmodified; if they are combined, possibility of each is changed to 0.25; further, if AC is added on, we set its possibility as 0.1 when AF and AFB do not happen. 

Specifically, in AF and AFB, linear pixel-wise augmentation, nonlinear pixel-wise augmentation and region-wise augmentation happen with a probability of $0.8$, $0.1$ and $0.1$, respectively. In AC, linear pixel-wise augmentation, nonlinear pixel-wise augmentation and region-wise augmentation happen with a probability of  $0.2$, $0.4$ and $0.4$, respectively. All the augmentations are randomly selected from the options list in the main text during each operation.

\section{Ablation Study on Strong Data Augmentation on the Composition-1k}
We report ablation study results on $\mathcal{SA}$ on the Composition-1k in Table.~\ref{tab:test_benchmark_dim}. In consistent with results in the main text, our $\mathcal{SA}$ produces comparable results on the synthetic benchmarks.

\begin{table}[!h]\small
    \centering
    \begin{tabular}{l | c c c c}
    \hline
    Method  & SAD & MSE & Grad & Conn \\
    \hline
         N3 & 25.86 & 0.0046 & 9.69 & 21.16 \\
         N3+AF & 25.86 & 0.0045 & 9.82 & 21.27 \\
         N3+AFB & 26.21 & 0.0048 & 9.91 & 21.43 \\
         N3+AF+AFB &  26.55 & 0.0050 & 10.45 & 21.93 \\
        N3+AF+AFB+AC  & 26.46 & 0.0049 & 9.98 & 21.72 \\
    \hline
    \end{tabular}
    \caption{Ablation on $\mathcal{SA}$ on the Composition-1k.}
    \label{tab:test_benchmark_dim}

\end{table}


\section{More Visual Results}
Here we show more visual results.  The visualized methods include IndexNet~\cite{lu2019indices}, CA~\cite{hou2019context}, GCA~\cite{li2020natural}, A$^2$U~\cite{dai2021learning}, SIM~\cite{sun2021semantic}, FBA~\cite{forte2020f} and our M7\textsuperscript{$\ddagger$}.

\begin{figure*}[!htb]
    \centering
    \includegraphics[height=22cm]{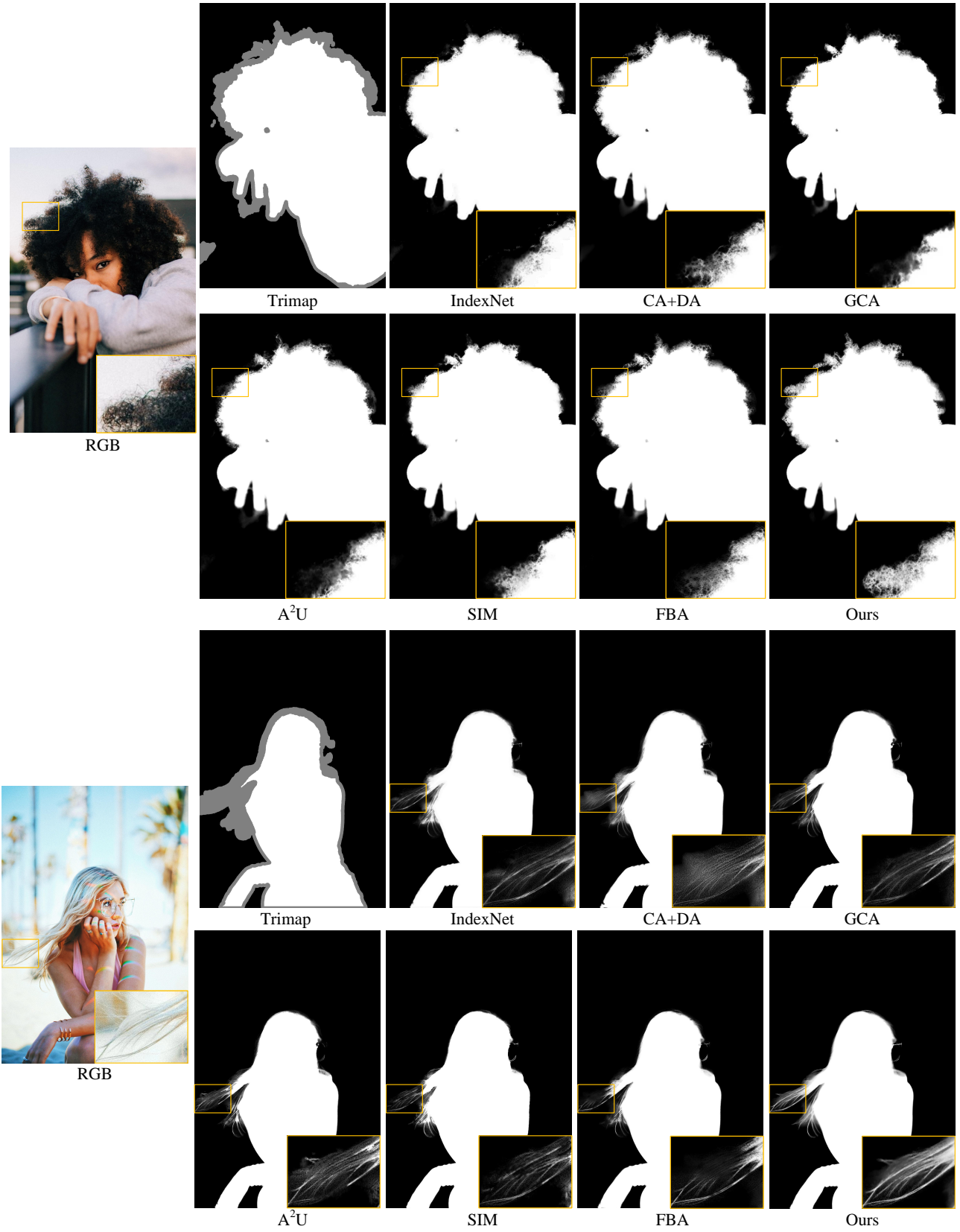}
    \caption{Visual results on real-world images. Best viewed by zooming in.}
    \label{fig:visual_rw}
\end{figure*}

\begin{figure*}[!htb]
    \centering
    \includegraphics[height=22cm]{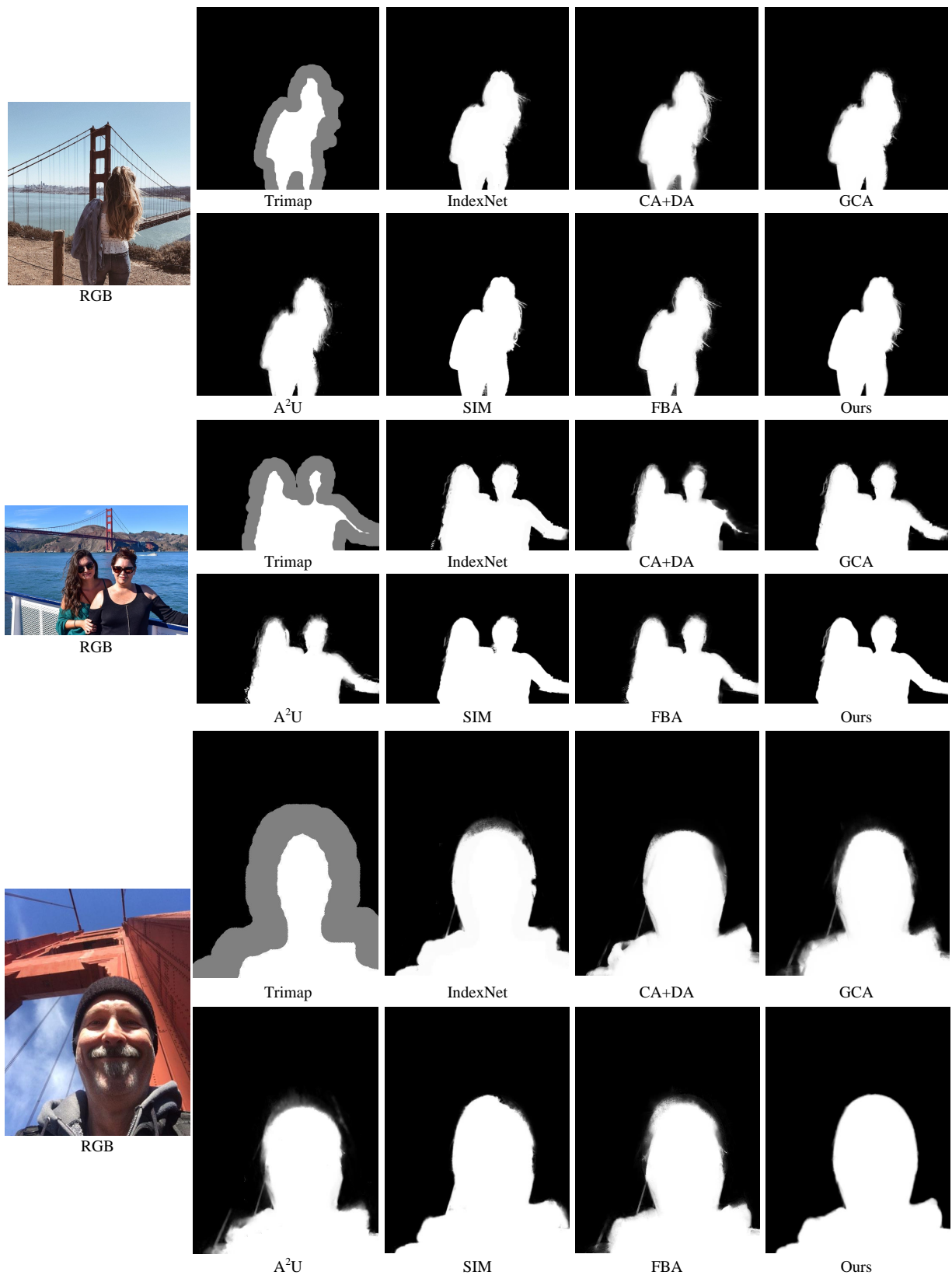}
    \caption{Visual results on real-world images with coarse trimaps.}
    \label{fig:visual_rw2}
\end{figure*}

Visual results on real-world images are present in Fig.~\ref{fig:visual_rw} and~\ref{fig:visual_rw2}, where Fig.~\ref{fig:visual_rw2} further shows results on coarse trimaps. Our method is more robust in these real-world test cases with coarse-to-fine trimaps.

\begin{figure*}[!htb]
    \centering
    \includegraphics[width=\linewidth]{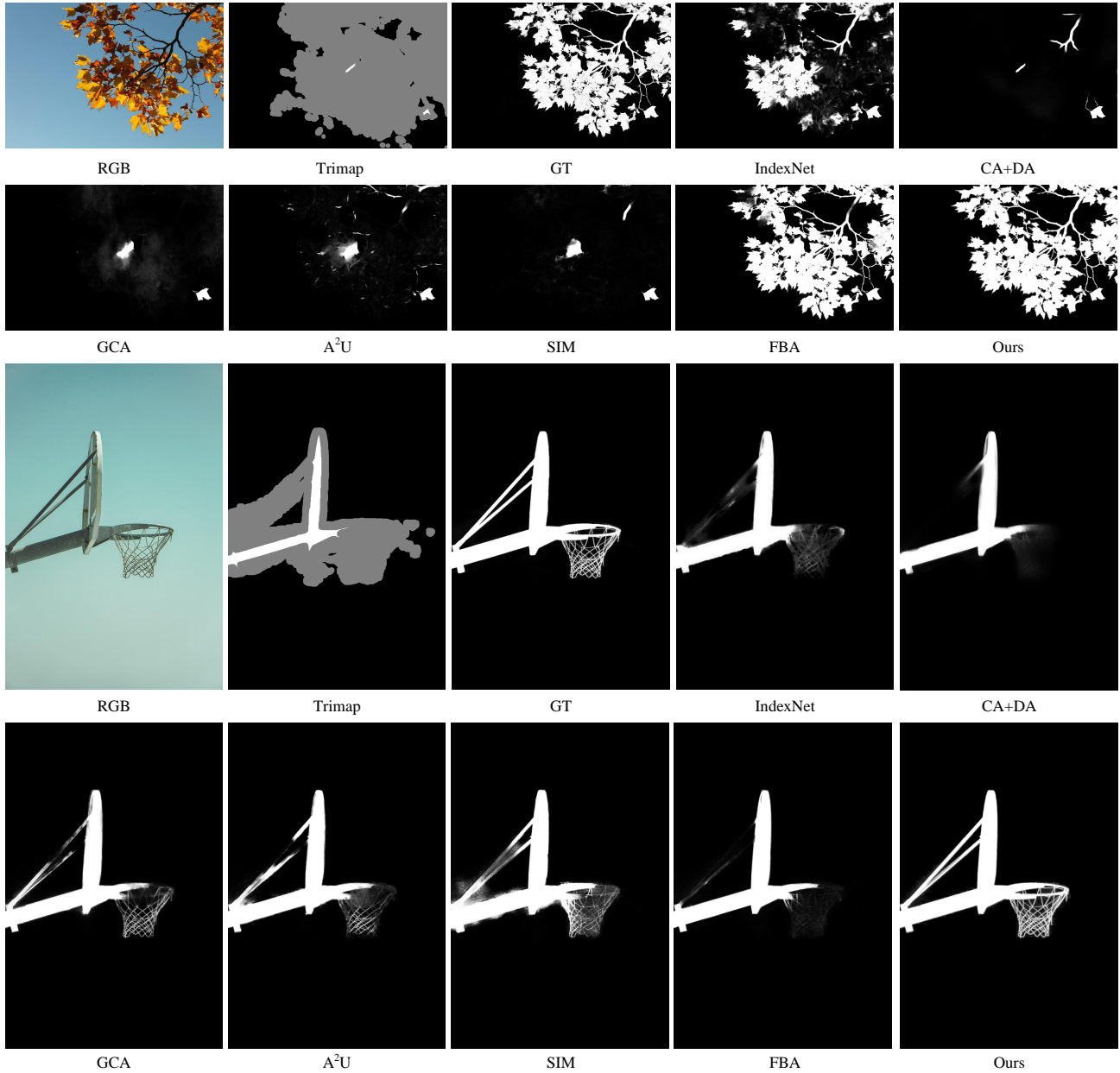}
    \caption{Visual results on the AIM-500.}
    \label{fig:visual_aim}
\end{figure*}

Visual results on the AIM-500~\cite{li2021deep} are exhibited in Fig.~\ref{fig:visual_aim}. Our method achieves better results on structures such as leaves and net.

\begin{figure*}[!htb]
    \centering
    \includegraphics[width=\linewidth]{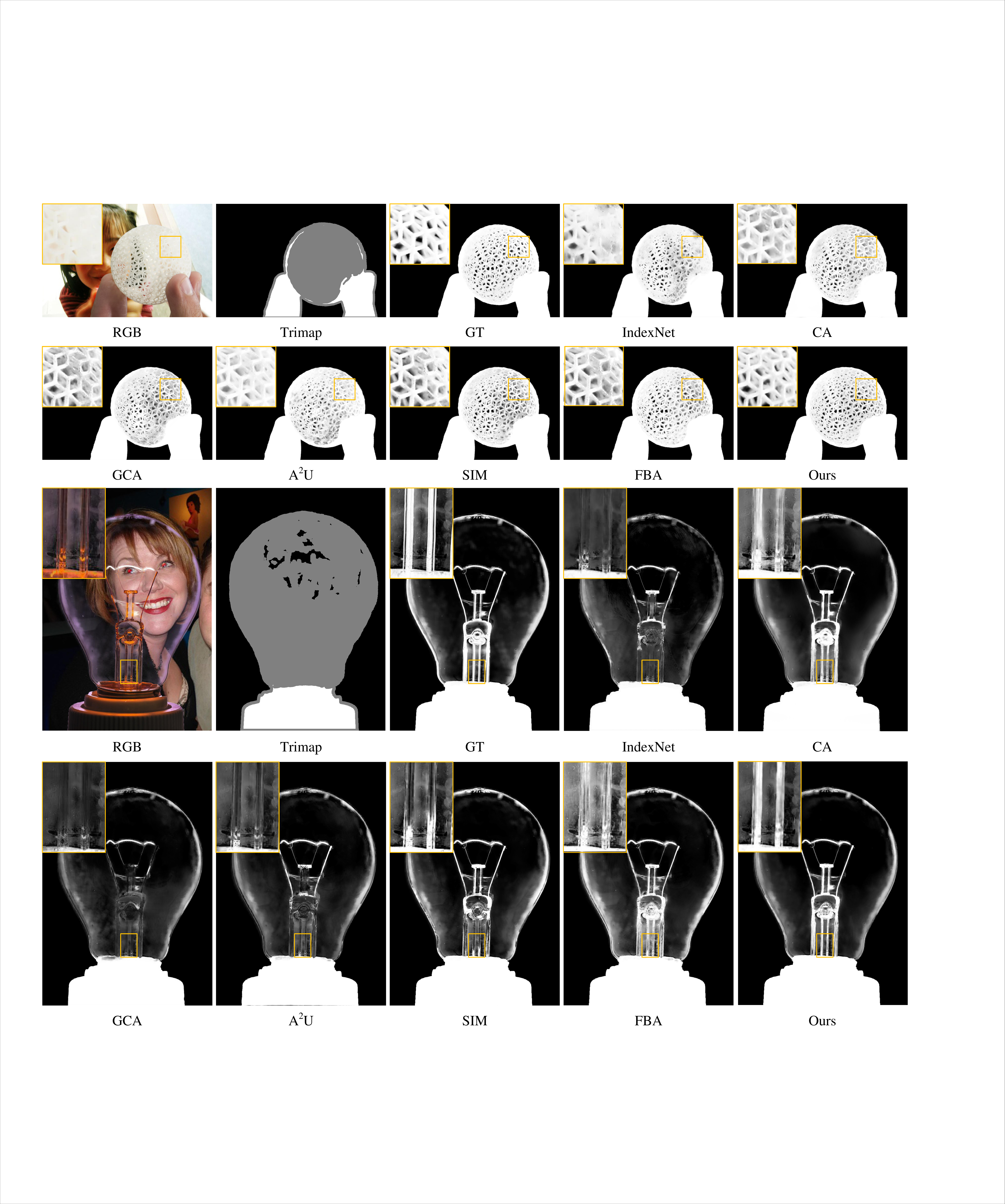}
    \caption{Visual results on the Composition-1k. Best viewed by zooming in.}
    \label{fig:visual_dim}
\end{figure*}

\begin{figure*}[!htb]
    \centering
    \includegraphics[width=\linewidth]{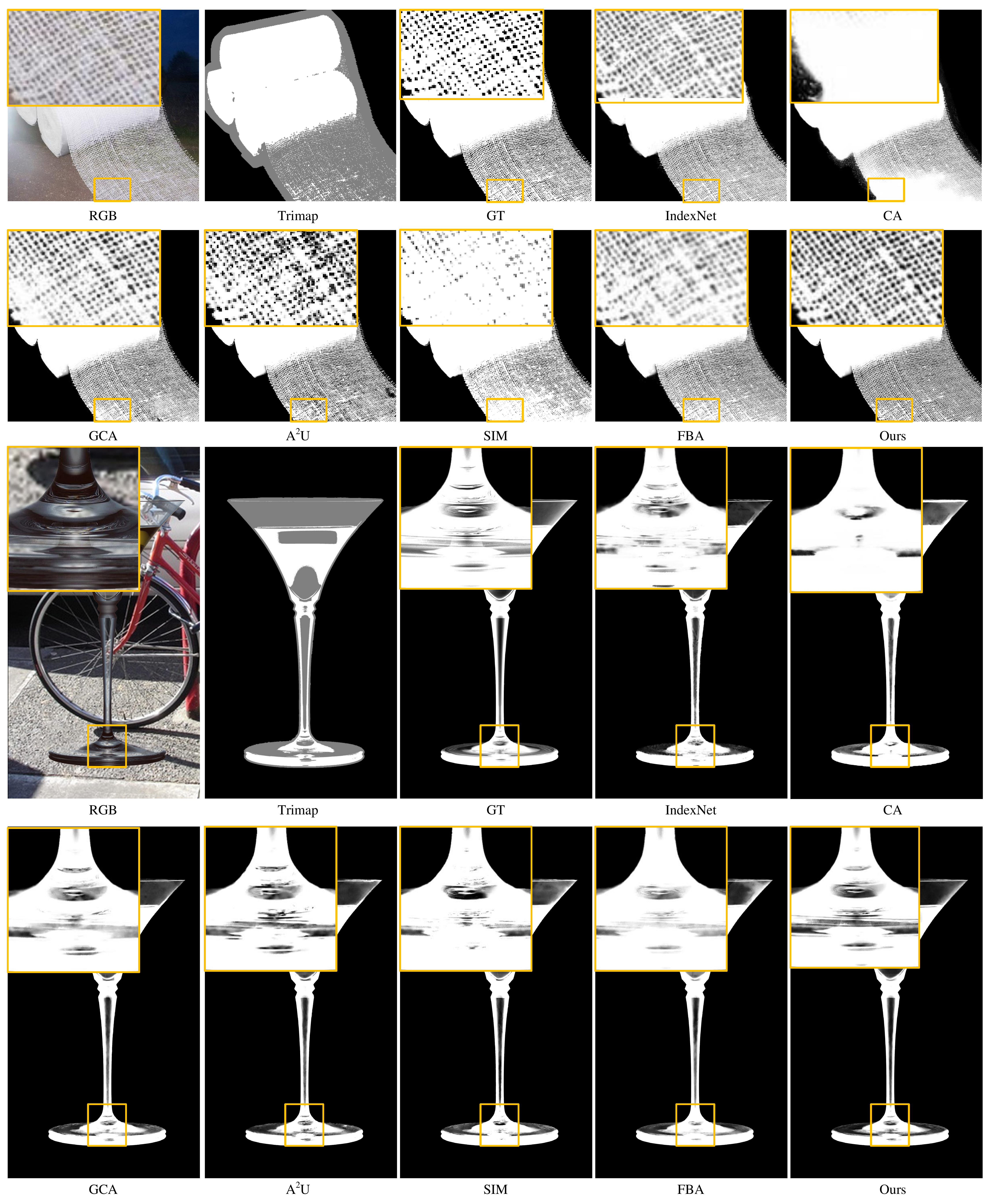}
    \caption{Visual results on the Distinction-646. Best viewed by zooming in.}
    \label{fig:visual_dis}
\end{figure*}

\begin{figure*}[!htb]
    \centering
    \includegraphics[width=\linewidth]{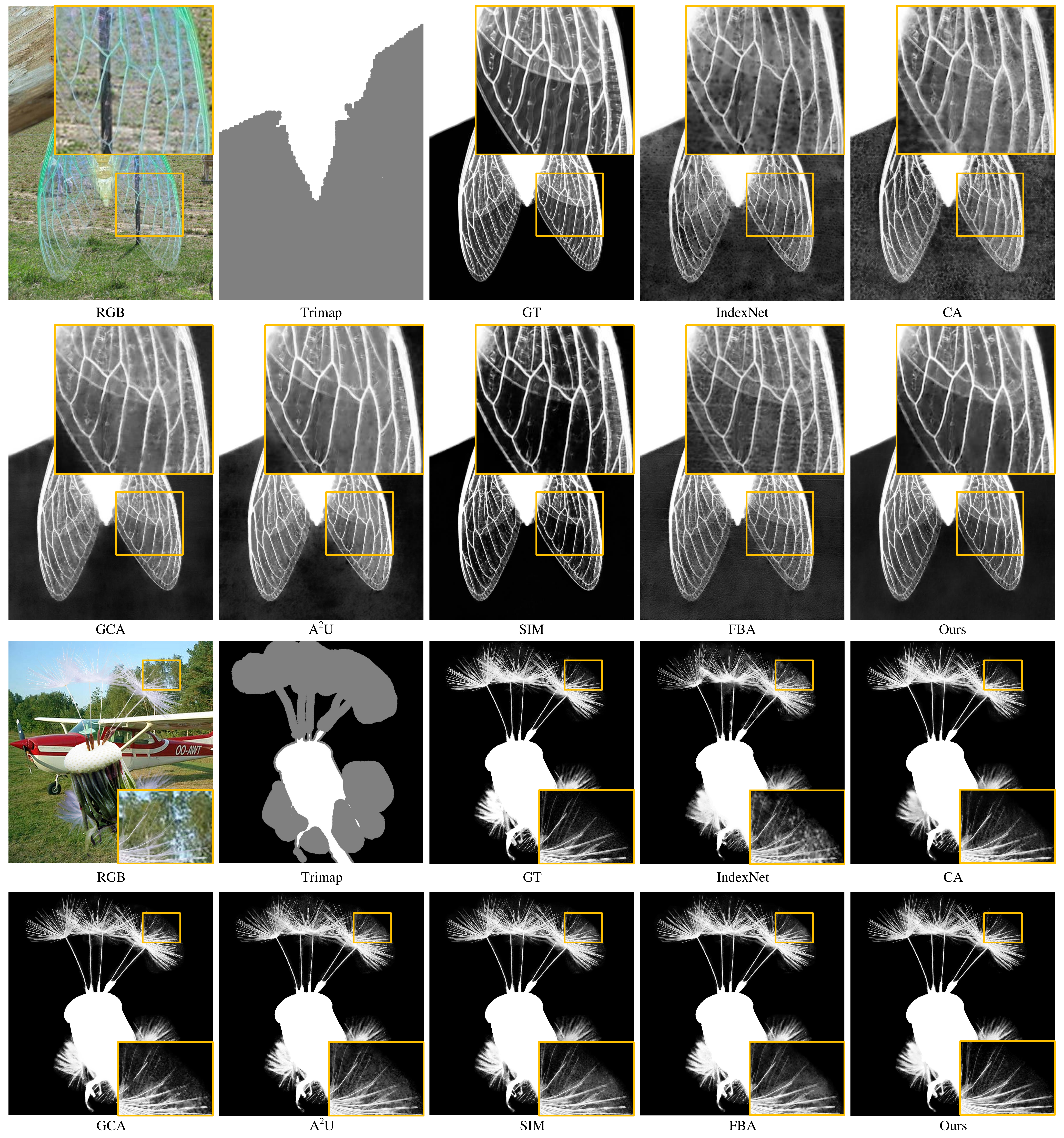}
    \caption{Visual results on the SIMD$_{our}$. Best viewed by zooming in.}
    \label{fig:visual_simd}
\end{figure*}

Visual results on the Composition-1k~\cite{xu2017deep}, Distinction-646~\cite{qiao2020attention} and SIMD$_{our}$~\cite{sun2021semantic} are displayed in Fig.\ref{fig:visual_dim}, ~\ref{fig:visual_dis} and~\ref{fig:visual_simd}, respectively. The top-performing methods in comparison all demonstrate appealing results on these synthetic benchmarks, but our method still performs better at background suppression (e.g. Fig.~\ref{fig:visual_simd}) and foreground structure modeling (e.g. Fig.~\ref{fig:visual_dim} and~\ref{fig:visual_dis}).

\section{Results on the $\tt alphamatting.com$}
We report results of M7\textsuperscript{$\ddagger$} on the $\tt alphamatting.com$ online benchmark in Table.~\ref{tab:alphamatting}. The methods in comparison are SIM~\cite{sun2021semantic}, A$^2$U~\cite{dai2021learning}, GCA~\cite{li2020natural}, CA~\cite{hou2019context}, IndexNet~\cite{lu2019indices}. There are only 8 test images in this online benchmark. It worth noting that, SIM is trained with the SIMD training set, which has $736$ foregrounds (including $360$ foregrounds from DIM) in the training set, while DIM only has $431$ foregrounds in the training set; GCA and A$^2$U retrain their models with the whole DIM dataset (including both training set and test set, $481$ foregrounds in total) for this benchmark. Our result is directly reported from M7\textsuperscript{$\ddagger$} trained with the DIM training set without using extra data or fine-tuning the model, but it still achieves top-performing ranks.

\begin{table*}[!hb]\scriptsize
    \centering
    \begin{tabular}{l c c c c c c c c c c c c c c c c}
    \hline
       Method  & \multicolumn{4}{c}{SAD} & \multicolumn{4}{c}{MSE} & \multicolumn{4}{c}{Grad} & \multicolumn{4}{c}{Conn} \\
         & overall & S & L & U &  overall & S & L & U &  overall & S & L & U & overall & S & L & U \\
    \hline
    Ours-M7\textsuperscript{$\ddagger$} & \underline{6.7} & \textbf{5.9} & \textbf{5.8} & \underline{8.5} & \textbf{6.8} & \textbf{5.9} & \textbf{5.5} & \underline{9.1} & \textbf{4.7} & \textbf{4.8} & \textbf{3.8} & \textbf{5.5} & \underline{12.4} & \underline{16.4} & \underline{13.7} & \textbf{7.6} \\
    SIM~\cite{sun2021semantic} & \textbf{6.5} & \underline{7} & \textbf{5.8} & \textbf{6.6} & \underline{7} & \underline{8.1} & \textbf{5.5} & \textbf{7.4} & \underline{6.9} & \underline{8.5} & \underline{5.9} & \underline{6.5} & \textbf{10} & \textbf{10} & \textbf{8.9} & \underline{11.3} \\
    A$^2$U~\cite{dai2021learning} & 13.3 & 12.3 & 10.6 & 17 & 15.5 & 13 & 12.6 & 20.8 & 12.3 & 11.3 & 9.4 & 16.1 & 27.3. & 30.1 & 28 & 24.3 \\
    GCA~\cite{li2020natural} & 14.5 & 15.3 & 12.4 & 16 & 15.3 & 15.1 & 14.5 & 16.4 & 13.7 & 13.6 & 12.5 & 15 & 22.5 & 26 & 20.1 & 21.4 \\
    CA~\cite{hou2019context} & 22.9 & 26.9 & 20.9 & 21 & 17.6 & 20.9 & 18.6 & 13.3 & 14.6 & 15.8 & 15.5 & 12.6 & 25.9 & 28 & 24.6 & 25 \\
    IndexNet~\cite{lu2019indices} & 19.4 & 21.5 & 18.1 & 18.6 & 22.9 & 25.3 & 21.5 & 22 & 18.6 & 17.3 & 17.3 & 21.4 & 25.5 & 24.1 & 26.4 & 26  \\
    \hline
    \end{tabular}
    \caption{Results on the $\tt alphamatting.com$ online benchmark.}
    \label{tab:alphamatting}
\end{table*}

\clearpage

\section{Failure Cases}
\begin{figure}[!h]
    \centering
    \includegraphics[width=\linewidth]{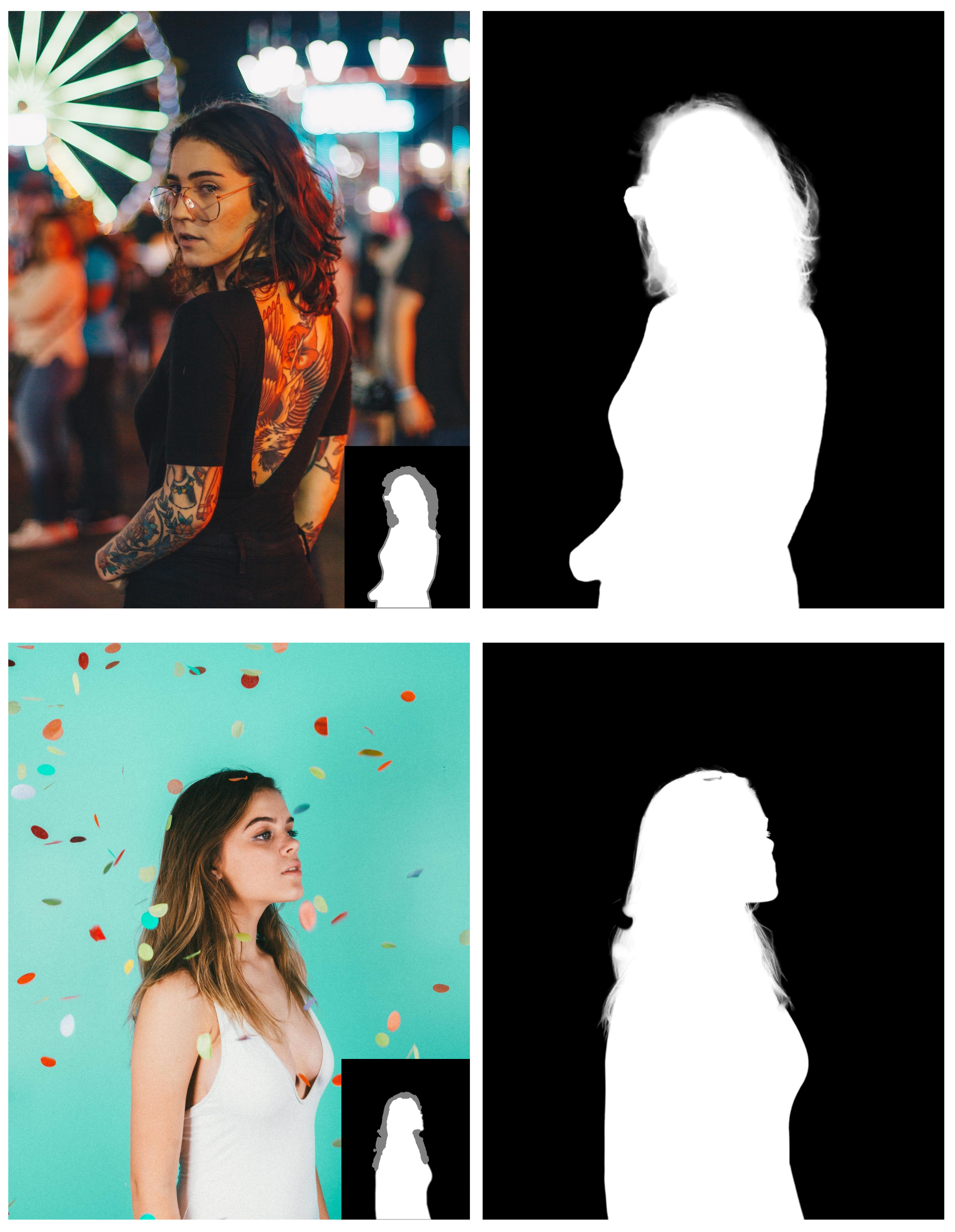}
    \caption{Failure cases.}
    \label{fig:failure_cases}
\end{figure}

Failure examples are visualized in Fig.~\ref{fig:failure_cases}. Our method may fail if there is strong light in the background or there are tiny objects overlapping with the foreground object. A possible solution is to learn the structure of the foreground objects. We leave it as a future work.

\end{document}